\documentclass[conference]{IEEEtran}
\IEEEoverridecommandlockouts
\usepackage{cite}
\usepackage{amsmath,amssymb,amsfonts}
\usepackage{textcomp}

\usepackage{marvosym}
\usepackage{url}
\usepackage{subfigure}
\usepackage{epsfig}
\usepackage{booktabs}
\usepackage{tabularx}
\usepackage{bm}
\usepackage{threeparttable}
\usepackage{tikz}
\tikzset{global scale/.style={
		scale=#1,
		every node/.append style={scale=#1}
	}
}
\usepackage{multirow}
\usepackage[linesnumbered,ruled,lined,commentsnumbered]{algorithm2e}

\newtheorem{problem}{Problem}
\newtheorem{definition}{Definition}
\def\BibTeX{{\rm B\kern-.05em{\sc i\kern-.025em b}\kern-.08em
    T\kern-.1667em\lower.7ex\hbox{E}\kern-.125emX}}
\usepackage{fancyhdr}

\begin{document}

\title{Higher-order Structure Based Anomaly Detection on Attributed Networks}

\author{\IEEEauthorblockN{1\textsuperscript{st} Xu Yuan}
\IEEEauthorblockA{\textit{School of Software Technology} \\
\textit{Dalian University of Technology}\\
Dalian, China \\
david@dlut.edu.cn}
\and
\IEEEauthorblockN{2\textsuperscript{nd} Na Zhou}
\IEEEauthorblockA{\textit{School of Software Technology} \\
\textit{Dalian University of Technology}\\
Dalian, China \\
zhouna824@mail.dlut.edu.cn}
\and
\IEEEauthorblockN{3\textsuperscript{rd} Shuo Yu\textsuperscript{(\Letter)}}
\IEEEauthorblockA{\textit{School of Software Technology} \\
	\textit{Dalian University of Technology}\\
	Dalian, China \\
	shuo.yu@ieee.org}
\and
\IEEEauthorblockN{4\textsuperscript{th} Huafei Huang}
\IEEEauthorblockA{\textit{School of Software Technology} \\
	\textit{Dalian University of Technology}\\
	Dalian, China \\
	hfhuang@mail.dlut.edu.cn}
\and
\IEEEauthorblockN{5\textsuperscript{th} Zhikui Chen}
\IEEEauthorblockA{\textit{School of Software Technology} \\
	\textit{Dalian University of Technology}\\
	Dalian, China \\
	zkchen@dlut.edu.cn}
\and
\IEEEauthorblockN{6\textsuperscript{th} Feng Xia}
\IEEEauthorblockA{\textit{School of Engineering, IT and Physical Sciences} \\
	\textit{Federation University Australia}\\
	Ballarat, Australia  \\
	f.xia@ieee.org}
}

\maketitle

\thispagestyle{fancy}
\fancyhead{}
\lhead{}
\lfoot{978-1-6654-3902-2/21/\$31.00~\copyright2021 IEEE}
\cfoot{}
\rfoot{}

\begin{abstract}
  Anomaly detection (such as telecom fraud detection and medical image detection) has attracted the increasing attention of people. The complex interaction between multiple entities widely exists in the network, which can reflect specific human behavior patterns. Such patterns can be modeled by higher-order network structures, thus benefiting anomaly detection on attributed networks. However, due to the lack of an effective mechanism in most existing graph learning methods, these complex interaction patterns fail to be applied in detecting anomalies, hindering the progress of anomaly detection to some extent. In order to address the aforementioned issue, we present a higher-order structure based anomaly detection (GUIDE) method. We exploit attribute autoencoder and structure autoencoder to reconstruct node attributes and higher-order structures, respectively. Moreover, we design a graph attention layer to evaluate the significance of neighbors to nodes through their higher-order structure differences. Finally, we leverage node attribute and higher-order structure reconstruction errors to find anomalies. Extensive experiments on five real-world datasets (i.e., ACM, Citation, Cora, DBLP, and Pubmed) are implemented to verify the effectiveness of GUIDE. Experimental results in terms of ROC-AUC, PR-AUC, and Recall@K show that GUIDE significantly outperforms the state-of-art methods.
\end{abstract}


\section{Introduction}
With the increase of the amount of network data, detecting anomalies from network data has become a significant research problem of urgent societal concerns \cite{Xia2021TAI}. Moreover, anomaly detection has a wide range of applications in real life \cite{Yu2020MIS}, such as financial fraud detection \cite{DBLP:journals/asc/VanhoeyveldMP20, DBLP:journals/dss/PourhabibiOKB20}, network intrusion detection \cite{DBLP:conf/iscc/ViegasSAO18}, web spam detection \cite{DBLP:journals/fgcs/MakkarK20}, and industrial anomaly detection \cite{DBLP:journals/sensors/RamotsoelaA018}. Therefore, the anomaly detection problem has attracted widespread attention from researchers \cite{Yu2020TCSSdetecting}.

\begin{figure}[t]
	\centering
	\subfigure[Attribute anomaly]{
		\includegraphics[scale=0.16]{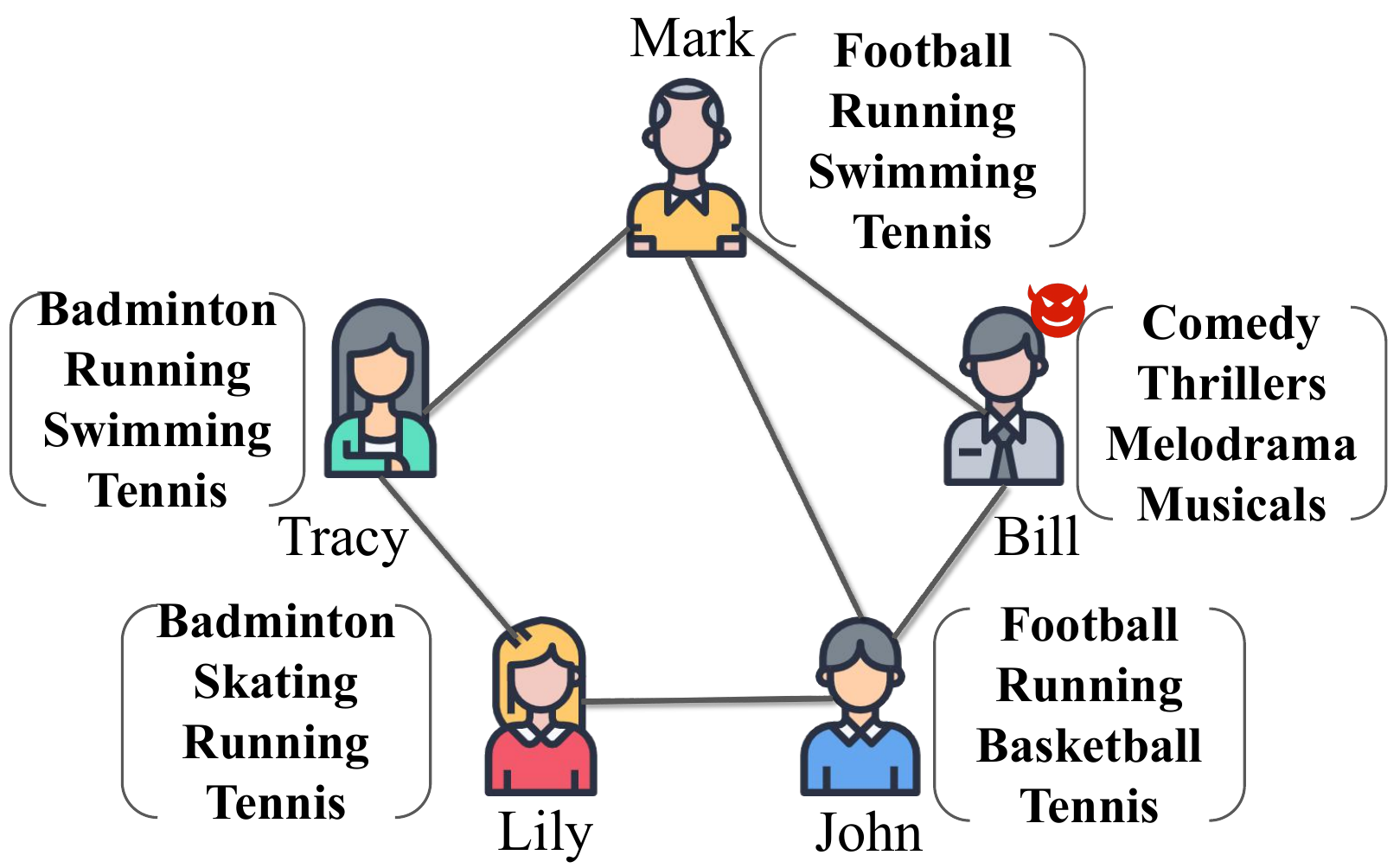}
	}
	\subfigure[Structural anomaly]{
		\includegraphics[scale=0.16]{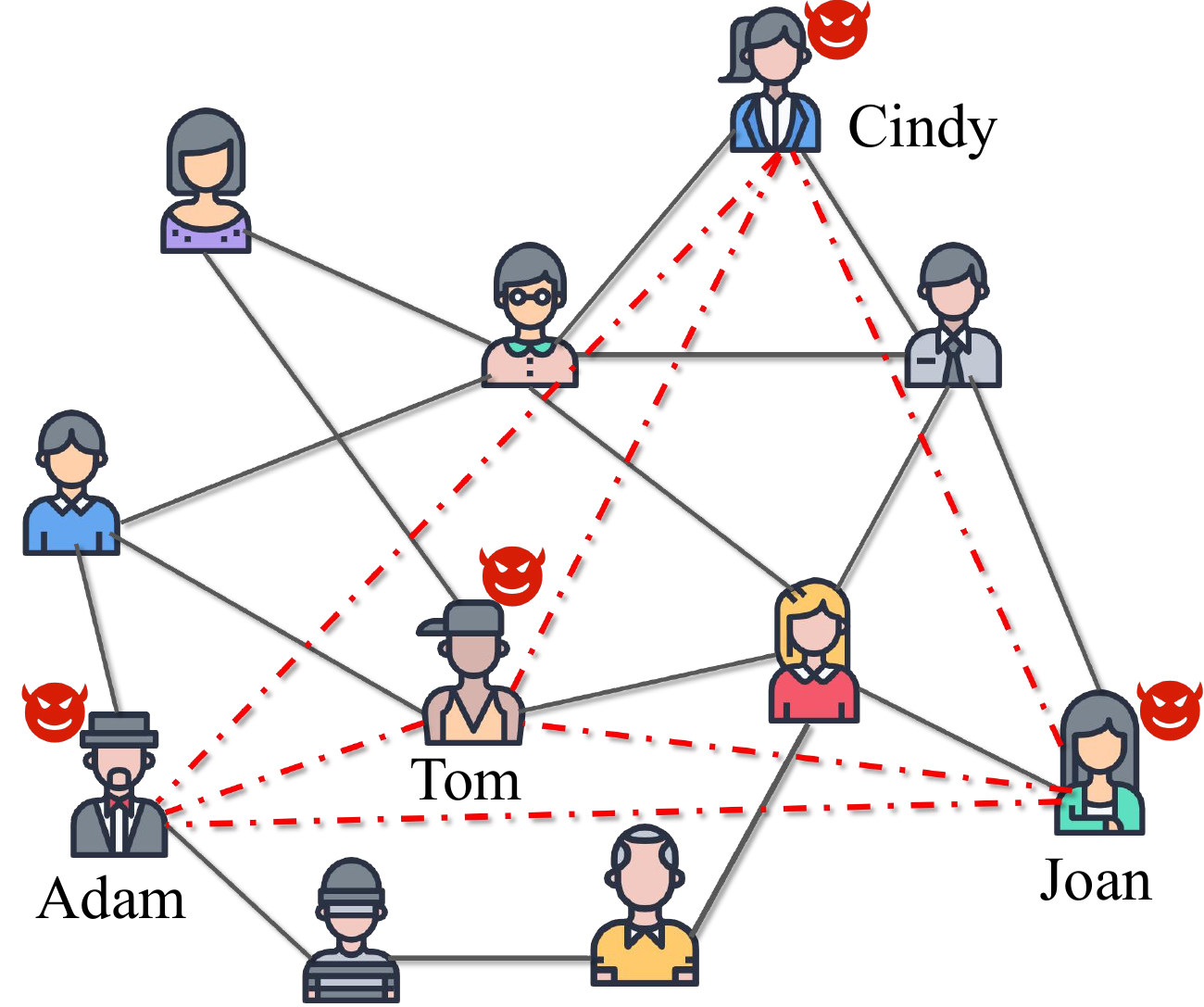}
	}
	
	\caption{The illustration about different types of anomalies on attributed networks.} 
	\label{fig:anomaly}
\end{figure}

Anomaly detection aims at finding the rare nodes whose behaviors are significantly different from other majority nodes. Furthermore, the abnormality of nodes on attributed networks depends on not only their abnormal situation of the network topology, but also the unusual condition of node attributes. Specifically, attribute abnormality mainly refers to the significant difference between the attributes of a node and its neighborhoods. For example, as shown in Fig \ref{fig:anomaly} (a), Tracy, Mark, Bill, John, and Lily are in the same interest group, but John's hobbies are quite different from the other members. The abnormal structure mostly refers to small groups that are far away but too closely connected. For instance, in Fig \ref{fig:anomaly} (b), Tom, Cindy, Joan, and Adam are members of a wire fraud organization. In order to facilitate the crime, they are closely related to each other. As it needs to simultaneously model the topological structure and node attributes, detecting anomalies of attributed networks is more challenging.

There have been many studies on detecting anomalies. Many studies attempt to find abnormal nodes through subspace selection of node feature \cite{DBLP:conf/icdm/SanchezMLKB13, DBLP:conf/ssdbm/SanchezMIB14}. Some other methods consider exploiting residual analysis to detect anomalies \cite{DBLP:conf/ijcai/LiDHL17, DBLP:conf/ijcai/PengLLLZ18}. However, these methods are based on shallow learning mechanisms and have certain limitations. For example, they can not model the complex interaction of network attributes and structures. With the increasing development of deep learning technologies,  the effectiveness of such kinds of methods has also been verified in addressing these problems~\cite{DBLP:conf/iscc/ZhuML20, DBLP:conf/wsdm/BandyopadhyayNV20, DBLP:conf/sdm/DingLBL19}. Deep neural networks are employed to encode attributed networks, and reconstruct the attributes and structures separately, which can utilize reconstruction errors to identify anomalies. However, previous studies still lack the ability of effectively utilizing complex interaction patterns among multiple entities to detect anomalies. When detecting anomalies, the significance of these complex interactions should be taken into consideration.

To address the above-mentioned problems, we present an unsupervised dual autoencoders framework, titled \textbf{GUIDE} (hi\textbf{\underline{G}}her-order str\textbf{\underline{U}}cture based anomaly detection on attr\textbf{\underline{I}}bute\textbf{\underline{D}} n\textbf{\underline{E}}tworks). Different from previous methods, we use the higher-order structures to model complex interaction patterns between multiple entities for anomaly detection in the network. To better learn higher-order network structures, we propose a graph node attention layer, which can learn different weights according to structural differences between the node and its neighbors. Specifically, we first encode attributes and higher-order structures of nodes to obtain the corresponding latent representation, and then exploit the decoder to reconstruct it. Finally, the reconstruction errors from both higher-order structure and attribute perspectives are used to detect anomalies of attributed networks. We summarize the main contributions of this paper as follows:

\begin{itemize}
	\item \textbf{Multiple Attributes-driven Anomaly Detection}: We propose a higher-order structure based anomaly detection method named GUIDE, which employs including node attributes and higher-order network structures to promote anomaly detection.
	\item \textbf{Higher-order Structure Attention Mechanism}: We design a higher-order structure attention mechanism, which utilizes structural differences between the node and its neighbors to generate attention weights. With this mechanism, our proposed GUIDE can better learn higher-order network structures.
	\item \textbf{Outperformance on Five Real-world Datasets}: Extensive experiments have been conducted on five real-world datasets, whose results show that GUIDE consistently outperforms all baseline methods significantly.
\end{itemize}

The rest of this paper is organized as follows. Section \ref{sec:rw} generalizes the related work. Section \ref{sec:definition} formally introduces the network motif and problem statement. In Section \ref{sec:model}, we introduce the design of the proposed GUIDE. Section \ref{sec:experiments} empirically evaluates GUIDE on five real-world datasets. Section \ref{sec:conclusion} concludes the whole paper.

\section{Related Work}\label{sec:rw}
\textbf{Anomaly Detection on Attributed Networks. }Compared with the plain (unlabeled) network, the attributed networks can model complex systems more effectively due to containing richer attribute information. Therefore, lots of researchers began to show interest in the problem of anomaly detection on attributed networks \cite{DBLP:conf/wsdm/DingLL19, DBLP:conf/ijcnn/ZhangYLPW19, DBLP:conf/dasfaa/XueLPLCL19}. For example, Perozzi et al. \cite{DBLP:conf/sdm/PerozziA16} leveraged attributes and network structure to quantify the quality of neighborhoods so that find anomalous neighborhoods on attributed networks. Li et al. \cite{DBLP:conf/ijcai/LiDHL17} found anomalous nodes by analyzing the residuals of attribute information and its coherence with network information. Moreover, Liu et al. \cite{DBLP:conf/ijcai/LiuHH17} introduced a novel anomaly detection model, learning simultaneously node attributes and structural information to effectively detect local anomalies on attributed networks. Peng et al. \cite{DBLP:conf/ijcai/PengLLLZ18} exploited CUR decomposition and residual analysis to filter out node attributes that are noisy and irrelevant, thereby avoiding their adverse effects for anomaly detection. Guti\'errez-G\'omez et al. \cite{gutierrez2020multi} explored all relevant contexts of anomalous nodes, and performed multiscalar anomaly detection on attributed networks. However, the above methods are limited by the shallow learning mechanism, so they cannot effectively learn the complex interactions between the node attribute and structure.

Driven by the great success of deep learning, a mass of studies have been devoted to exploiting deep neural networks to detect anomalous nodes on attributed networks. For instance, Ding et al. \cite{DBLP:conf/sdm/DingLBL19} constructed a deep autoencoder using graph convolutional neural networks, and evaluated the abnormality of nodes through the reconstruction errors of node attributes and structure. Li et al. \cite{DBLP:conf/cikm/LiHLDZ19} utilized Laplacian sharpening to magnify the distance between representations of anomalous nodes and normal nodes, making it easier to find anomalies. Ding et al. \cite{DBLP:conf/ijcai/DingLAL20} presented an adversarial graph differential network, utilizing generative adversarial ideas to detect anomalies on the attribute network, which can be naturally extended to newly observed data. Furthermore, Chen et al. \cite{DBLP:conf/cikm/ChenLWDLB20} came up with a generative adversarial attributed networks anomaly detection model. By obtaining the sample reconstruction error generated by the generator and the discriminant loss of real node pairs, this algorithm can predict effectively abnormal nodes. Despite the above approaches achieving superior performance over other shallow methods, they cannot effectively utilize complex interaction patterns between multiple entities to detect anomalies.

\textbf{Higher-order Network Representation Learning. }The complex real network contains a wealth of higher-order structures (i.e., motifs), which reflect the internal relationships of nodes in the network \cite{DBLP:conf/jcdl/XuYSRLP020}. Multiple studies have confirmed that it is effective to consider higher-order structures in network representation learning \cite{DBLP:conf/www/RossiZA18, DBLP:conf/www/PiaoZXCL21, DBLP:conf/asunam/YangLZ018}. The framework for learning higher-order network embedding was proposed by Rossi et al. \cite{DBLP:conf/www/RossiAK18}, aiming at utilizing various motif-based matrix formulations to learn effectively network embedding. Lee et al. \cite{DBLP:conf/cikm/LeeRKKKR19} exploited a motif-based attention mechanism to learn higher-order interactions between nodes with their neighbours. In \cite{DBLP:conf/cikm/Yu0XCL20}, Yu et al. chose proper motifs to strengthen the multivariate relationships, and improve the learning effect of higher-order graph representation. In addition, Xu et al. \cite{DBLP:conf/jcdl/XuYSRLP020} aggregated the higher-order structure features and attribute features of nodes to obtain the final network embedding, and demonstrate superior performance in the node classification task. Liu et al. \cite{DBLP:conf/www/LiuHYD21} simultaneously modeled the local higher-order structures and temporal evolution to learn node representation for dynamic attributed networks. Nevertheless, all the aforementioned methods focus on network representation learning, it is still not clear how to effectively utilize higher-order structures for anomaly detection.

\begin{figure}[htbp]
	\centering
	\includegraphics[width=0.39\textwidth]{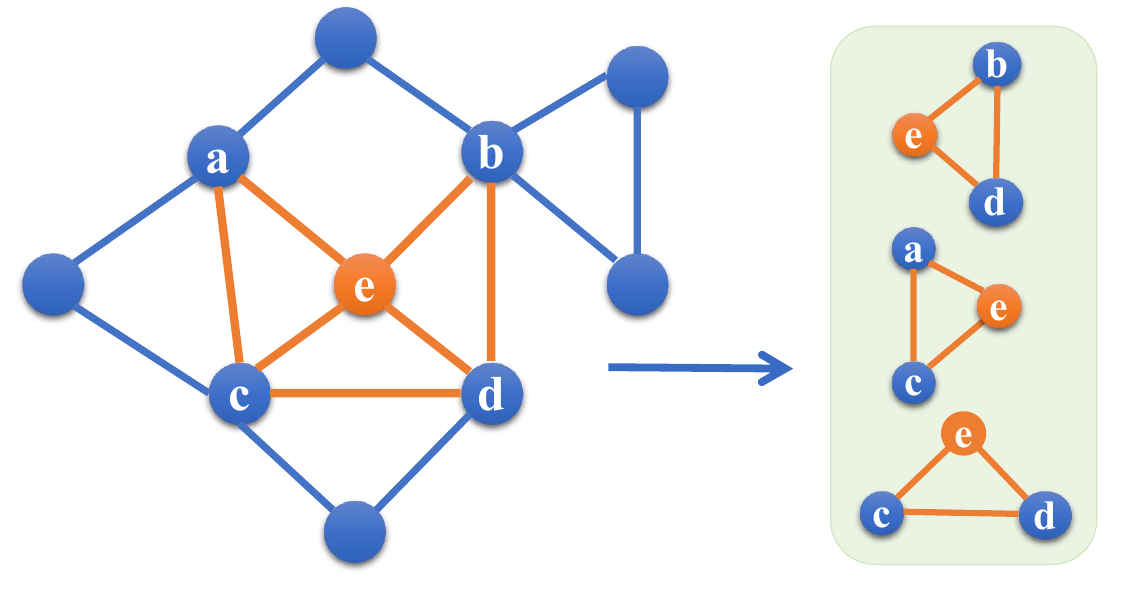}
	\caption{A node motif degree calculation of M31 (For node e, it contains three M31 motifs, i.e., ebd, eac, ecd).}
	\label{fig:motif}
\end{figure}

\section{Preliminaries}\label{sec:definition}

\subsection{Network Motif}
Network motifs \cite{milo2002network} refer to special subgraph structures
frequently appearing in the network. Paranjape et al. \cite{paranjape2017motifs} studied the timing network and found that network motifs help understand the crucial structure of the network. In addtion, network motifs \cite{Yu2020CSRmotif} have specific practical significance. For instance, a three-order triangle motif can describe the collaboration relationship of three scholars in the academic network. Therefore, we can utilize the network motif to effectively model complex interaction patterns between multiple entities in the network.

\begin{figure}[t]
	\centering
	\subfigure[M31]{
		\includegraphics[scale=0.38]{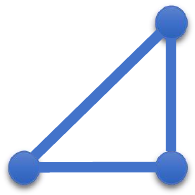}
	}
	\subfigure[M32]{
		\includegraphics[scale=0.38]{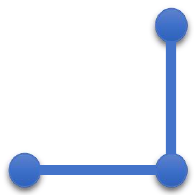}
	}
	\subfigure[M41]{
		\includegraphics[scale=0.38]{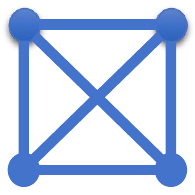}
	}
	\subfigure[M42]{
		\includegraphics[scale=0.38]{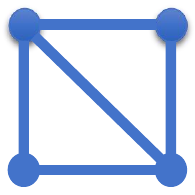}
	}
	\subfigure[M43]{
		\includegraphics[scale=0.38]{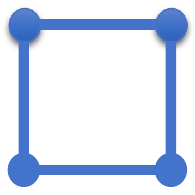}
	}
	
	\caption{Network motifs used in this paper.} 
	\label{fig:motifs}
\end{figure}

Because network motifs comprised of five or more nodes are so complex and numerous that it is difficult to deal with them. In this paper, we adopt network motifs comprised of three or four nodes to analyze the network. In Fig \ref{fig:motifs}, we list the motif types used in this paper.

We employ the node motif degree proposed by Yu et al. \cite{DBLP:conf/cikm/Yu0XCL20} to represent the higher-order structures of nodes in this paper. Specifically, the node motif degree is defined as follows.
\begin{definition}
	\textbf{Node Motif Degree (NMD):} 
	For the graph $\mathcal{G}=(\mathcal{V}, \mathcal{E})$, a node $i \in V$, the node motif degree of $i$ is $NMD(i)$, which represents node i is involved in the number of the motif $M$. 	
\end{definition}
As shown in Fig \ref{fig:motif}, the node $e$ contains the number of \bm{$M31$} motifs is 3.

\begin{figure*}[!t]
	\centerline{\includegraphics[scale=0.42]{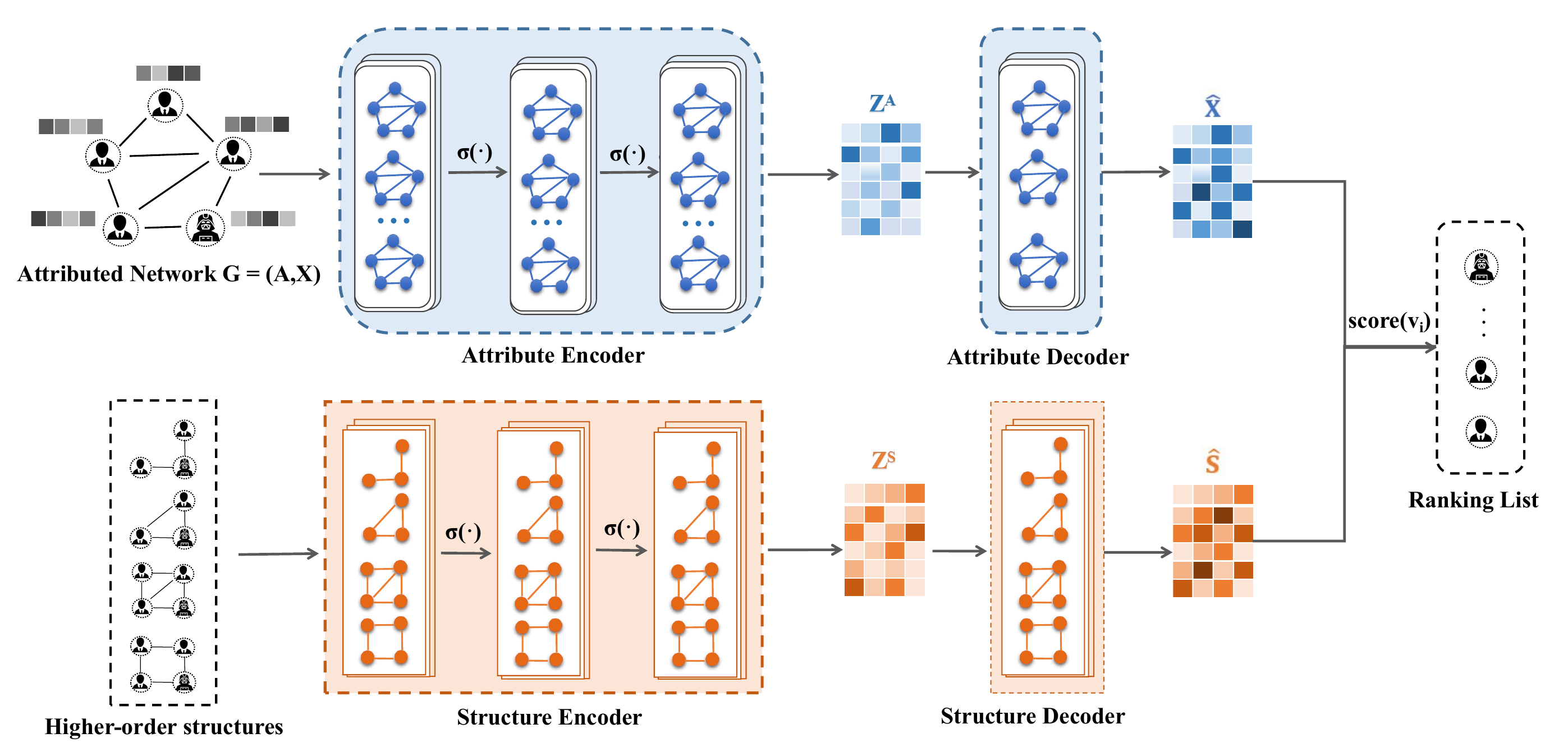}}
	\caption{The framework of the proposed GUIDE.}
	\label{fig:framework}
\end{figure*}
\subsection{Problem Definition}
In this paper, we use bold lowercase letters (e.g., $\mathbf{x}$) and bold uppercase letters (e.g., $\mathbf{X}$), to denote vectors and matrices, respectively. Besides, we use calligraphic fonts (e.g., $\mathcal{V}$) to represent sets. The $i^{th}$ row of a matrix $\mathbf{X}$ is denoted by $\mathbf{x}_i$ and ${(i,j)}^{th}$ element of matrix $\mathbf{X}$ is denoted by $\mathbf{X}_{i,j} $. The notations mainly used in this paper are summarized in Table \ref{table:notation}.  

\begin{definition}
	\textbf{Attributed Networks:} 
	Give an attributed network $\mathcal{G}=(\mathbf{A}, \mathbf{X})$, where $\mathbf{A} \in \mathbb{R}^{n \times n}$ is the adjacency matrix, $\mathbf{X} \in \mathbb{R}^{n \times d}$ is the attribute matrix. The $i^{th}$ row vector $\mathbf{x}_i \in \mathbb{R}^{d}$ of the attribute matrix $\mathbf{X}$ represents the $i^{th}$ node’s attribute vector. Besides, $\mathbf{A}_{i,j} = 1$ if there is an edge between node $i$ and node $j$, otherwise $\mathbf{A}_{i,j} = 0$.
\end{definition}
\begin{definition}
	\textbf{Structure Matrix:}
	The higher-order structures of $\mathcal{G}$ can be represented by a structure matrix $\mathbf{S}$. The $i^{th}$ row vector $\mathbf{s}_i \in \mathbb{R}^{m}$ of the structure matrix $\mathbf{S}$ represents the $i$ th node’s structure vector, which is composed of the node motif degrees of \bm{$M31$}, \bm{$M32$}, \bm{$M41$}, \bm{$M42$}, \bm{$M43$}, and the original degree of the node.
\end{definition}

\begin{problem}
	\textbf{Anomaly Detection}. Given an attributed network $\mathcal{G} = \{\mathbf{A},\mathbf{X}\}$, the task is to rank all the nodes according to their anomalous scores ($score(v_i)$), and the node that are significantly different from the majority nodes ($\ge0.9n$) should obtain higher score and be ranked higher than other nodes.
\end{problem}
Next, we will describe in detail the GUIDE model which models node attributes and higher-order structures jointly to detect anomalies of the network.

\section{ The Proposed Model - GUIDE}\label{sec:model}
In this section, we introduce the proposed model GUIDE in detail. The framework of our approach is illustrated in Fig. \ref{fig:framework}. We design two essential components for GUIDE: attribute autoencoder and structure autoencoder, which are respectively responsible for reconstructing node attributes and higher-order structures. Then, we use reconstruction errors of node attributes and higher-order structures to calculate anomaly scores of nodes and rank them. Finally, anomalies in the network can be found by the ranking list.
\begin{table}[t]
	\caption{Notations and explanations related to GUIDE.} 
	\small
	\centering
	\begin{tabular}{ p{46 pt}<{\centering} | p{175 pt}}  
		\toprule[1.0pt]
		\textbf{Notation} & \textbf{Explanation}  \\
		\cmidrule{1-2}
		$\mathcal{G}=(\mathbf{A}, \mathbf{X})$ & An attributed network \\
		$\mathbf{A} \in \mathbb{R}^{n \times n}$ & The adjacency matrix of $\mathcal{G}$\\
		$\mathbf{X} \in \mathbb{R}^{n \times d}$ & The attribute matrix of $\mathcal{G}$ \\
		$\mathbf{S} \in \mathbb{R}^{n \times m}$ & The higher-order structures matrix of $\mathcal{G}$ \\
		$\mathbf{x}_i \in \mathbb{R}^{d}$ & The attribute vector of the $i^{th}$ node in $\mathcal{G}$ \\
		$\mathbf{s}_i \in \mathbb{R}^{m}$ & The higher-order structure vector of the $i^{th}$ node in $\mathcal{G}$ \\
		$n$ & The number of nodes in $\mathcal{G}$ \\
		$d$ & The dimension of attributes in $\mathcal{G}$ \\
		$m$ & The dimension of higher-order structures in $\mathcal{G}$ \\
		\cmidrule{1-2}
		$|\cdot|$ & The number of elements of a set \\
		$\sigma(\cdot)$ & The non-linear activation function \\
		$\mathbf{X}^{T}$ & The transpose of a matrix  $\mathbf{X}$\\
		$\|\cdot\|_F$ & The Frobenius norm of a matrix or vector \\
		$\|\cdot\|_2$ & The $\ell_2$-norm of a matrix or vector \\
		\bottomrule[1.0pt]
	\end{tabular}
	\label{table:notation}
\end{table}

\subsection{Attribute Autoencoder}

In this part, we aim at designing an effective autoencoder to reconstruct node attributes, thereby catching attribute anomalies. Formally, an autoencoder network layer is described as:
\begin{equation}
\label{eq:GCN_definition}
\mathbf{H}^{(l + 1)} = \sigma\left( {\overset{\sim}{\mathbf{D}}}^{- \frac{1}{2}}\overset{\sim}{\mathbf{A}}{\overset{\sim}{\mathbf{D}}}^{- \frac{1}{2}}\mathbf{H}^{(l)}\mathbf{W}^{(l)} \right),
\end{equation}
where $\mathbf{H}^{(l)}$ is the latent representation of the input in layer $l$, $\widetilde{\mathbf{A}}=\mathbf{A}+\mathbf{I}$ denotes the adjacency matrix of the attributed network $\mathcal{G}$ with added self-connections and $\widetilde{\mathbf{D}}_{i,i}=\sum_{j}\widetilde{\mathbf{A}}_{i,j}$. $\sigma(\cdot)$ represents a non-linear activation function. We adopt $\mathbf{Relu}$ as activation function in this paper. $\mathbf{W}^{(l)}$ is the trainable weight matrix. We let $\overset{-}{\mathbf{A}} = {\overset{\sim}{\mathbf{D}}}^{- \frac{1}{2}}\overset{\sim}{\mathbf{A}}{\overset{\sim}{\mathbf{D}}}^{- \frac{1}{2}}$. Then the autoencoder network layer formula can be abbreviated as:
\begin{equation}
\label{eq:GCN_abbr_definition}
\mathbf{H}^{(l + 1)} = f_{{\mathit{Re}l}u}(\overset{-}{\mathbf{A}} \mathbf{H}^{(l)} \mathbf{W}^{(l)}).
\end{equation}
In this paper, we encode the attributed networks using three autoencoder network layers, and the attribute matrix $\mathbf{X} \in \mathbb{R}^{n \times d}$ is regarded as the original input features:
\begin{equation}
\mathbf{H}^{(0)} = \mathbf{X},
\end{equation}
Therefore, the attribute encoder can be expressed as:
\begin{equation}
\label{eq: GCN_encode1}
\mathbf{H}^{(1)} = f_{{\mathit{Re}l}u}(\overset{-}{\mathbf{A}} \mathbf{X} \mathbf{W}^{(0)}),
\end{equation}
\begin{equation}
\label{eq:GCN_encode2}
\mathbf{H}^{(2)} = f_{{\mathit{Re}l}u}(\overset{-}{\mathbf{A}} \mathbf{H}^{(1)} \mathbf{W}^{(1)}),
\end{equation}
\begin{equation}
\label{eq:GCN_encode3}
\mathbf{Z}^{A} = \mathbf{H}^{(3)} = f_{{\mathit{Re}l}u}(\overset{-}{\mathbf{A}} \mathbf{H}^{(2)} \mathbf{W}^{(2)}).
\end{equation}
After using three autoencoder network layers, the encoder compresses the node attributes and network topology to get a low-dimensional latent representation $\mathbf{Z}^{A}$ of attributed networks. To reconstruct the node attributes, we exploit an autoencoder network layer to approximate the original attributes of nodes, which can be expressed as follows: 
\begin{equation}
\label{eq:GCN_decode}
\mathbf{\hat{X}} = f_{{\mathit{Re}l}u}( \overset{-}{\mathbf{A}} \mathbf{Z}^{A} \mathbf{W}^{(3)} ).
\end{equation}
Here, $\mathbf{\hat{X}}$ represents the reconstructive attribute matrix. Therefore, the attribute reconstruction loss can be calculated:
\begin{equation}
\label{eq:Attribute reconstruction loss}
\mathbf{R}_{A} = \mathbf{X} - \mathbf{\hat{X}}.
\end{equation}

\subsection{Structure Autoencoder}

Considering the importance of higher-order structures to the anomaly detection on attributed networks, in this part, we use structure autoencoder to reconstruct the higher-order structures of nodes. And the calculated structure reconstruction loss can capture structural anomalies. Specifically, nodes with abnormal structures are usually too closely connected to some nodes on attributed networks. Their higher-order structures are very different from normal nodes, and cannot be reconstructed well.

Therefore, inspired by \cite{DBLP:conf/ijcai/DingLAL20}, we designed a graph node attention network (GNA) to encode the higher-order structures of the node. It can better learn the structural difference between the node and its neighbors by utilizing the higher-order structures attention mechanism and help identify structural anomalies. Specifically, a graph node attention layer can learn the representation of node $i$ in layer $l$ , which can be formulated as follows: 
\begin{equation}
\label{eq :GNA_definition}
\mathbf{h}_{i}^{(l + 1)} = \sigma\left( \mathbf{W}_{1}\mathbf{h}_{i}^{(l)} + {\sum\limits_{j \in N{(i)} \cup \{ i\}}{\alpha_{ij}\mathbf{W}_{2}\mathbf{h}_{j}^{(l)}}} \right),
\end{equation}
where, $\mathbf{h}_{i}^{(l)}$, $\mathbf{h}_{j}^{(l)}\in \mathbb{R}^{F}$ is the input representation of node $i$ and node $j$, respectively. $N(i)$ represents the neighborhood of node $i$. $\mathbf{h}_{i}^{(l+1)}\in \mathbb{R}^{{F}^{'}}$ is the output representation of node $i$. $\mathbf{W}_{1}$,$\mathbf{W}_{2}\in \mathbb{R}^{{F}^{'} \times F}$ are two trainable weight matrices, and $\sigma(\cdot)$ denotes the $\mathbf{Relu}$ activation function. To determine the importance of different nodes, we calculate the normalized attention coefficient $\alpha_{ij}$ by:
\begin{equation}
\label{eq :attention}
\alpha_{ij} = \frac{{\mathit{\exp}(}\mathbf{a}^{T}\mathbf{W}_{2}( {\mathbf{h}_{i}}^{(l)} - {\mathbf{h}_{j}}^{(l)} ))}{\sum_{k \in N(i) \cup \{ i\}}{{\mathit{\exp}(}\mathbf{a}^{T}\mathbf{W}_{2}( {\mathbf{h}_{i}}^{(l)} - {\mathbf{h}_{k}}^{(l)} ))}}.
\end{equation}

where, ${\mathbf{h}_{i}}^{(l)} - {\mathbf{h}_{j}}^{(l)}$ represents the higher-order structures difference between node $i$ and node $j$. $\mathbf{a}\in \mathbb{R}^{{F}^{'}}$ is a parametrized weight vector. We generate the attentional weights based on the higher-order structures differences to facilitate the characterization of structural abnormality of node.

Similar to the attribute encoder, we use the three graph node attention layers to encode the higher-order structures of the node to obtain the corresponding latent representation $\mathbf{Z}^{S}$. In order to reconstruct structure matrix $\mathbf{S}$, we use another graph node attention layer to approximate the original higher-order structures of the node, which is expressed as follows:
\begin{equation}
\label{eq :structure_decode}
\mathbf{\hat{S}} = \mathbf{graph\_ node\_ atten}( \mathbf{Z}^{S} ),
\end{equation}

Here, $\mathbf{graph\_node\_atten(\cdot)}$ represents the operation of the graph node attention network described earlier. $\mathbf{\hat{S}}$ represents the reconstructive structure matrix. Therefore, the structure reconstruction loss can be calculated:
\begin{equation}
\label{eq :Structure reconstruction loss}
\mathbf{R}_{S} = \mathbf{S} - \mathbf{\hat{S}}.
\end{equation}
We can detect anomalous nodes of the network from the perspective of the higher-order structure. 
\begin{algorithm}[t]
	\SetKwInOut{Param}{Param}\SetKwInOut{Input}{Input}\SetKwInOut{Output}{Output}\SetKwInOut{Setup}{Setup}
	\SetKwFunction{InteractionAgg}{AGG} 
	\SetKwFunction{Eval}{DETECT}
	\SetKwFunction{Union}{UNION}
	\SetKwFunction{Update}{F-FAC} 
	\Input{Attributed network $\mathcal{G}=(\mathbf{A}, \mathbf{X})$, Training epochs  $Epoch_{AE}$}
	\Output{Well-trained GCN-AE and GNA-AE}
	$i\leftarrow 0$\;
	\While{$i<Epoch_{AE}$}{
		Compute the reconstructed node attributes $\hat{X}$ via Eq. (\ref{eq:GCN_decode})\;
		Compute the reconstructed higher-order network structures $\hat{S}$ via Eq. (\ref{eq :structure_decode})\;
		Update GCN-AE and GNA-AE with the loss function Eq. (\ref{eq :Total loss})\;
	}
	\caption{The training process of GUIDE}
	\label{alg:GUIDE}
\end{algorithm}

\subsection{Loss Function and Anomaly Detection}
To jointly learn the reconstruction errors, we aim to minimize the loss function of both network higher-order structures and node attribute: 
\begin{equation}
\label{eq :Total loss}
\begin{aligned}
\mathcal{L} &= ( {1 - \alpha})\mathbf{R}_{S} + \alpha \mathbf{R}_{A} \\
&= ( 1 - \alpha ) || \mathbf{S} - \mathbf{\hat{S}} | |_{F}^{2} + \alpha || \mathbf{X} - \mathbf{\hat{X}} | |_{F}^{2}.
\end{aligned}
\end{equation}
where $\alpha$ is a balance parameter which controls the training weight of higher-order structures reconstruction errors and attribute reconstruction errors. 
\begin{figure*}[t]
	\centering
	\subfigure[ACM]{
		\includegraphics[scale=0.28]{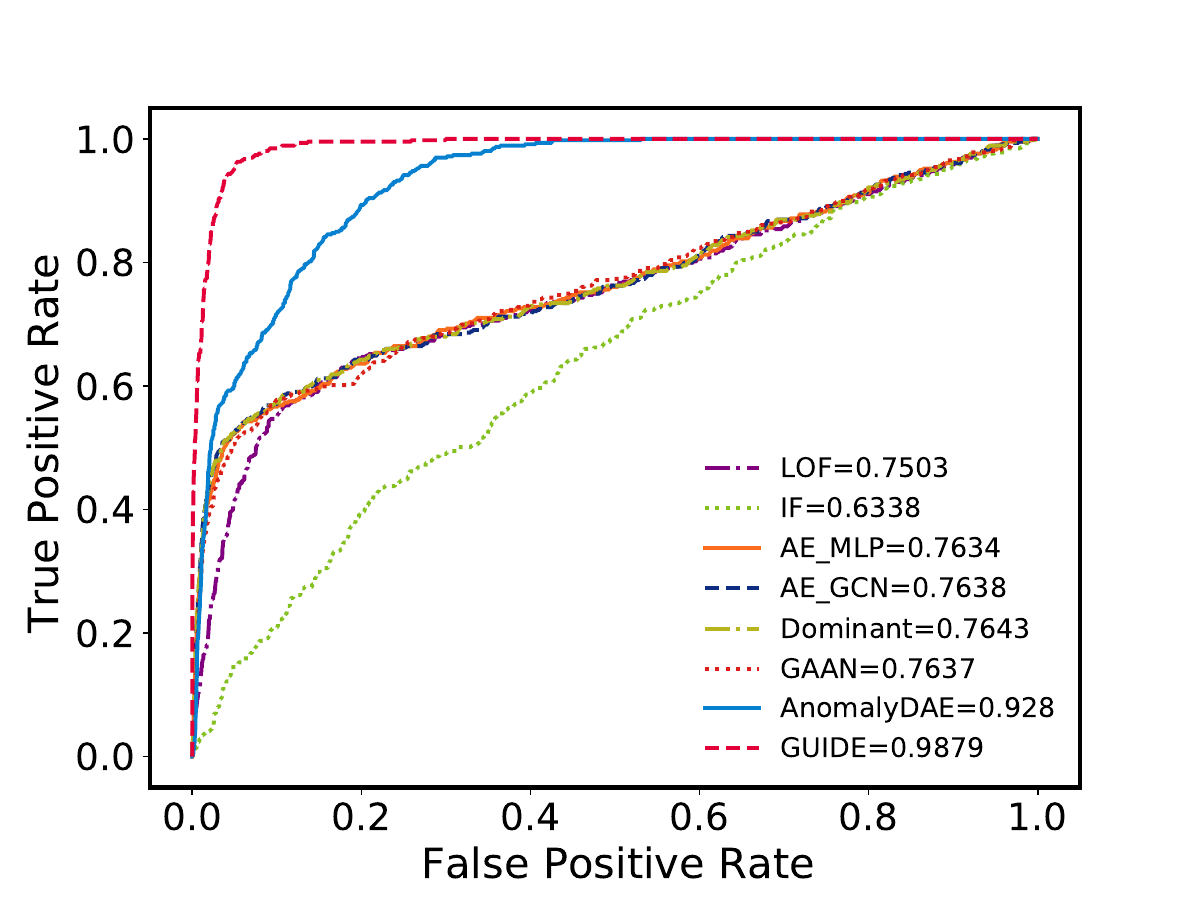}
	}
	\subfigure[Citation]{
		\includegraphics[scale=0.28]{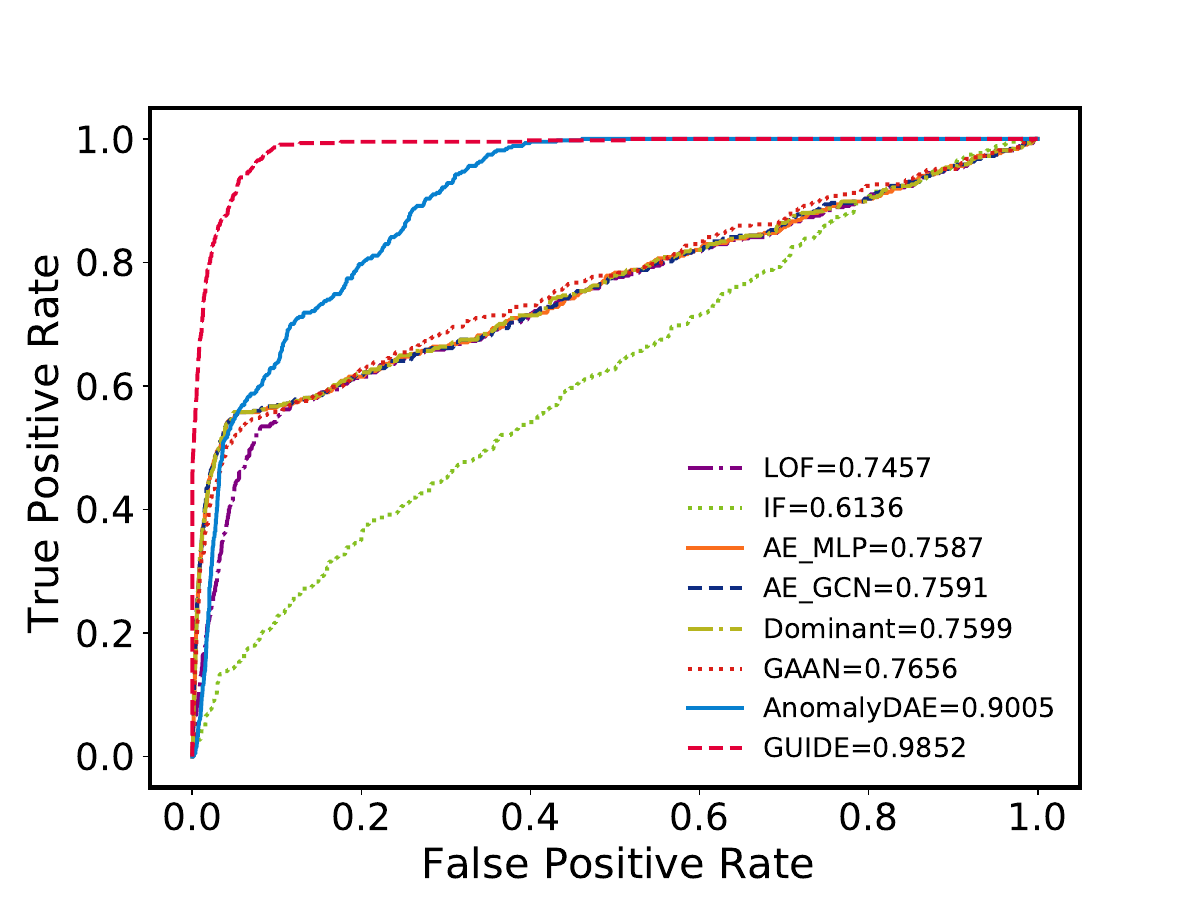}
	}
	\subfigure[Cora]{
		\includegraphics[scale=0.28]{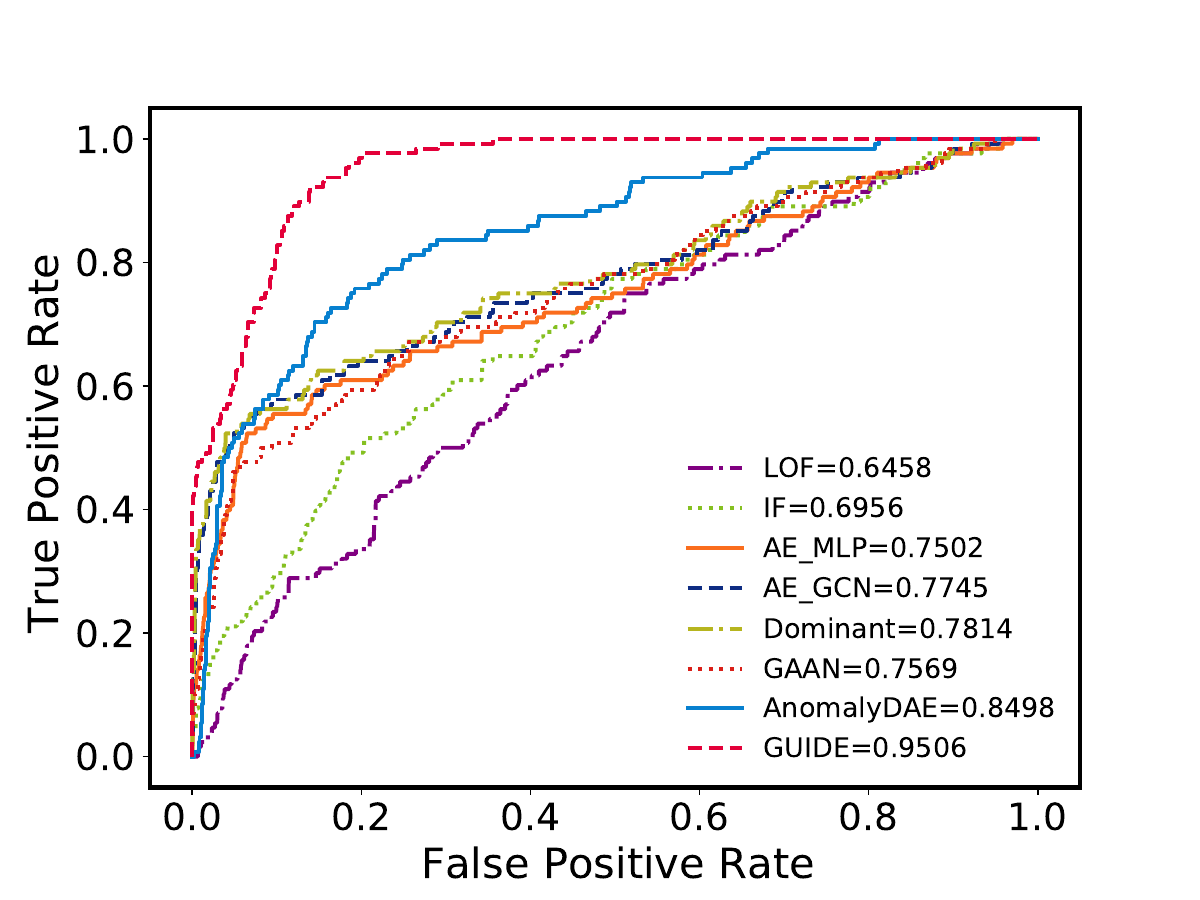}
	}
	\subfigure[DBLP]{
		\includegraphics[scale=0.28]{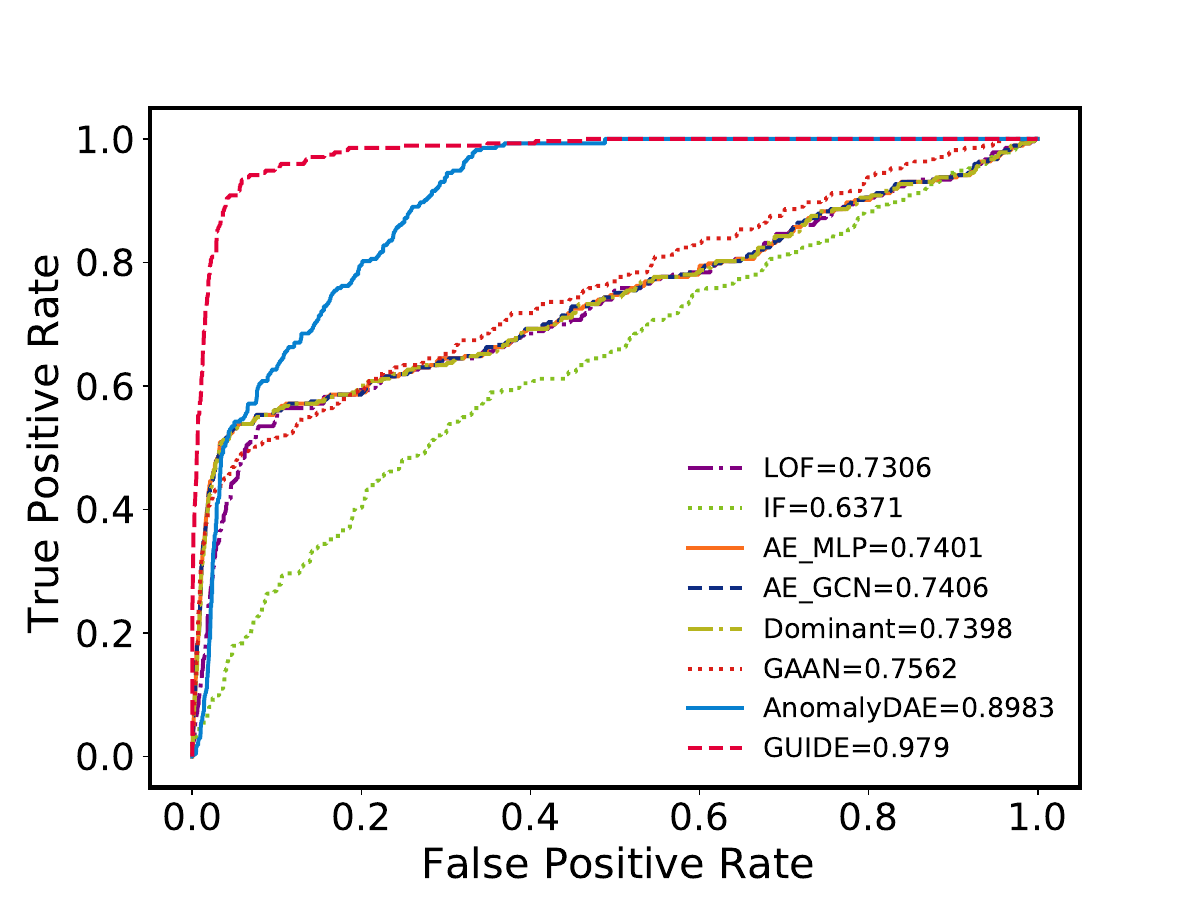}
	}
	\subfigure[Pubmed]{
		\includegraphics[scale=0.28]{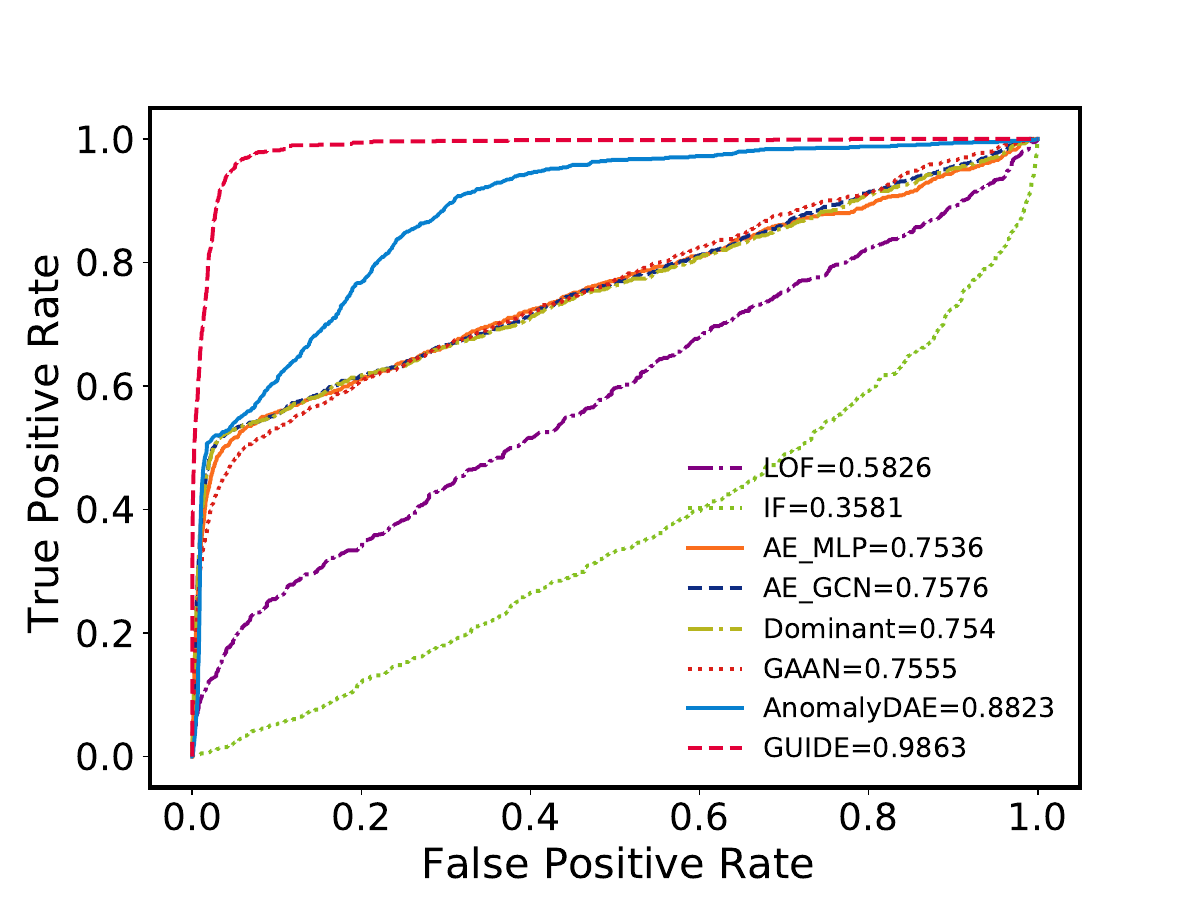}
	}
	\caption{ ROC curves and AUC scores of all methods on five real-world datasets.} 
	\label{fig:ROC AUC}
\end{figure*}
We can utilize the reconstruction error to assess the anomaly degree of nodes. Specifically, the attributes or higher-order structures of a node cannot be reconstructed well, indicating that its behavior pattern deviates from the majority of other nodes, and the probability of it being an anomalous node is higher. Thus, we can use the reconstruction error from both higher-order structures and node attribute perspective to calculate the anomaly score of each node:
\begin{equation}
\label{eq :Anomaly_score}
score( \mathbf{v}_{i} ) = \left( 1 - \alpha \right) | | \mathbf{s}_{i} - {\mathbf{\hat{s}}}_{i}{||}_{F}^{2} + \alpha || \mathbf{x}_{i} - \mathbf{{\hat{x}}}_{i} ||_{F}^{2}.
\end{equation}

Note that the node with a higher score has a higher probability of being an abnormal node. So we can rank all nodes by their anomaly scores. And the detailed model training process is shown in Algorithm \ref{alg:GUIDE}.

\section{Experiments}\label{sec:experiments}
In this section, we perform extensive experiments on five real-world datasets to confirm the effectiveness of the GUIDE model.

\subsection{Datasets}
To comprehensively evaluate the GUIDE model, we choose five real-world datasets that have been used in many previous studies \cite{DBLP:conf/sdm/DingLBL19, DBLP:conf/iccS/ZhuML20, DBLP:conf/aaai/BandyopadhyayLM19} in our experiments: 

\begin{table}[!t]
	\caption{Details of the five real-world datasets with injected anomalies.}
	\centering
	\resizebox{0.46\textwidth}{!}{
		\begin{threeparttable}{
				\begin{tabular}{cccccc}
					\toprule
					\textbf{Network Name} & \textbf{ACM}     & \textbf{Citation} & \textbf{Cora} & \textbf{DBLP} & \textbf{Pubmed}    \\
					\midrule
					\textbf{\#nodes}      & 9,360          &8,935                  & 2,708         &5,484	  & 19,717         \\
					\textbf{\#edges}      & 15,556		  &15,098                 & 5,278         &8,117	  &44,338       \\
					\textbf{\#attributes} & 6,775		  &6,775                  & 1,433           & 6,775	&500          \\
					\textbf{\#anomalies}      & 5$\%$     & 5$\%$                & 5$\%$          &5$\%$     & 5$\%$           \\
					
					\midrule
					\textbf{\#M31}   & 3,898		          & 3,716                & 1,630          &1,788	   &12,550     \\
					\textbf{\#M32}       & 66,214		  & 97,839               & 47,411         &41,536	&661,782    \\
					\textbf{\#M41}        & 678		      & 723                 & 220           &414	&3,291  \\
					\textbf{\#M42}        & 6,575		  & 8,174                & 2,468         &3,680	   &53,407     \\
					\textbf{\#M43}        & 7,002		  & 5,825                & 1,536         &3,177	  &100,440    \\
					\bottomrule
			\end{tabular}}
	\end{threeparttable}}
	
	\label{Table:DS}
\end{table}
\begin{figure*}[t]
	\centering
	\subfigure[ACM]{
		\includegraphics[scale=0.28]{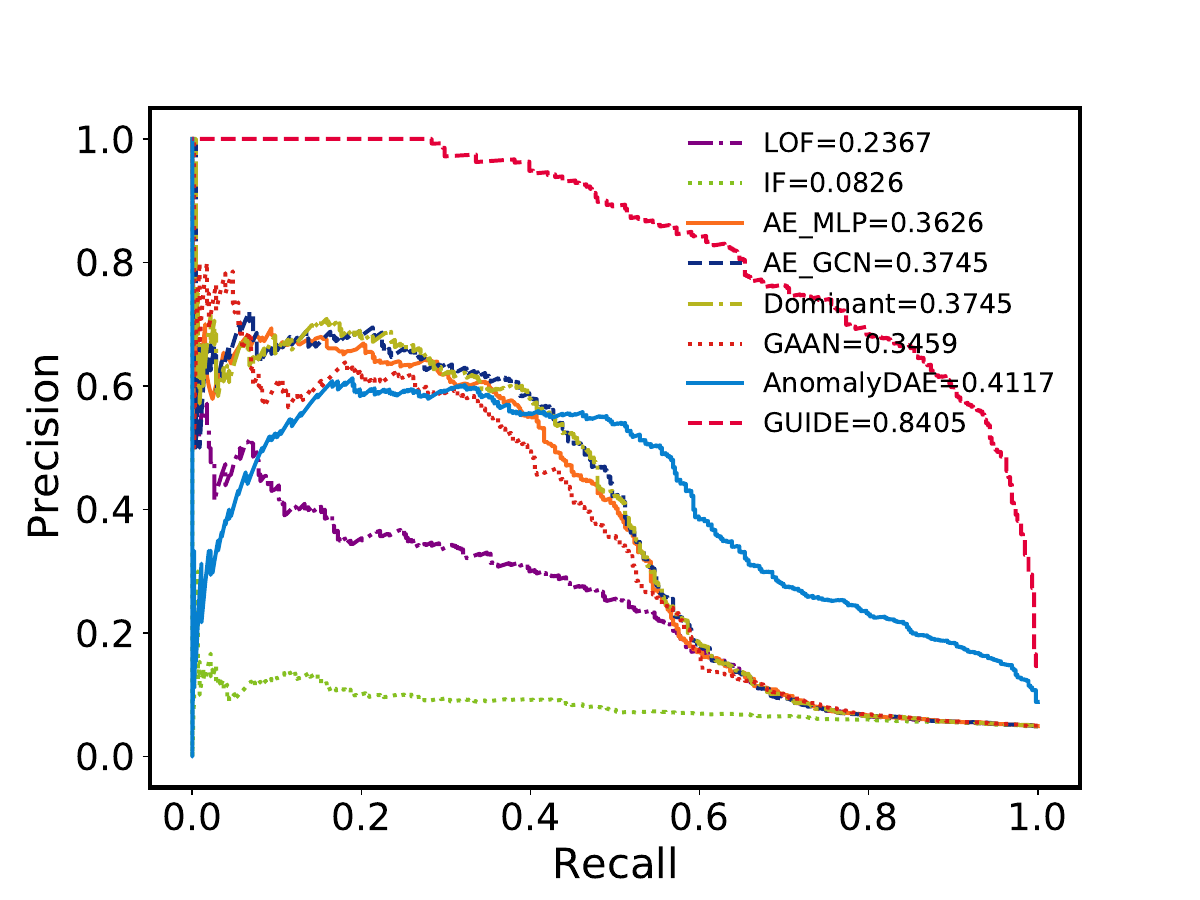}
	}
	\subfigure[Citation]{
		\includegraphics[scale=0.28]{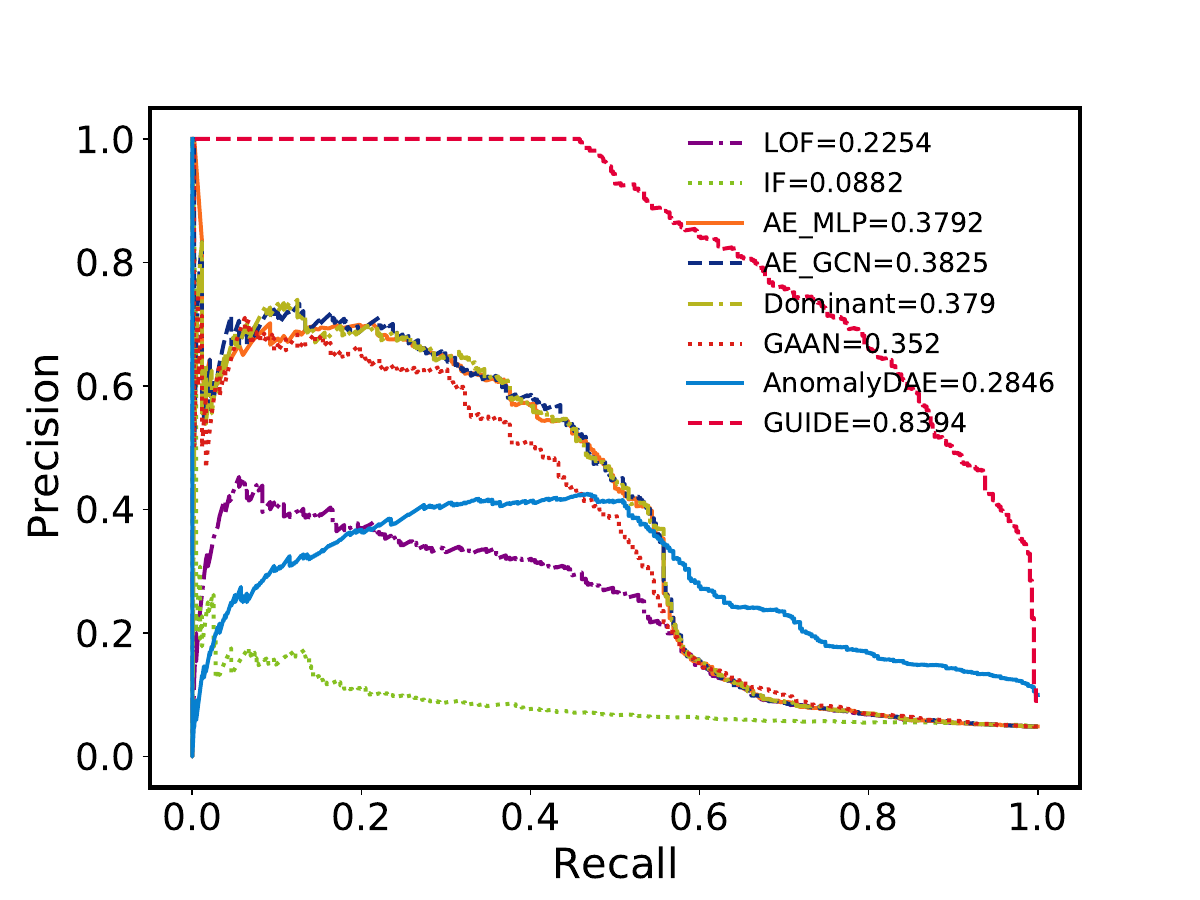}
	}
	\subfigure[Cora]{
		\includegraphics[scale=0.28]{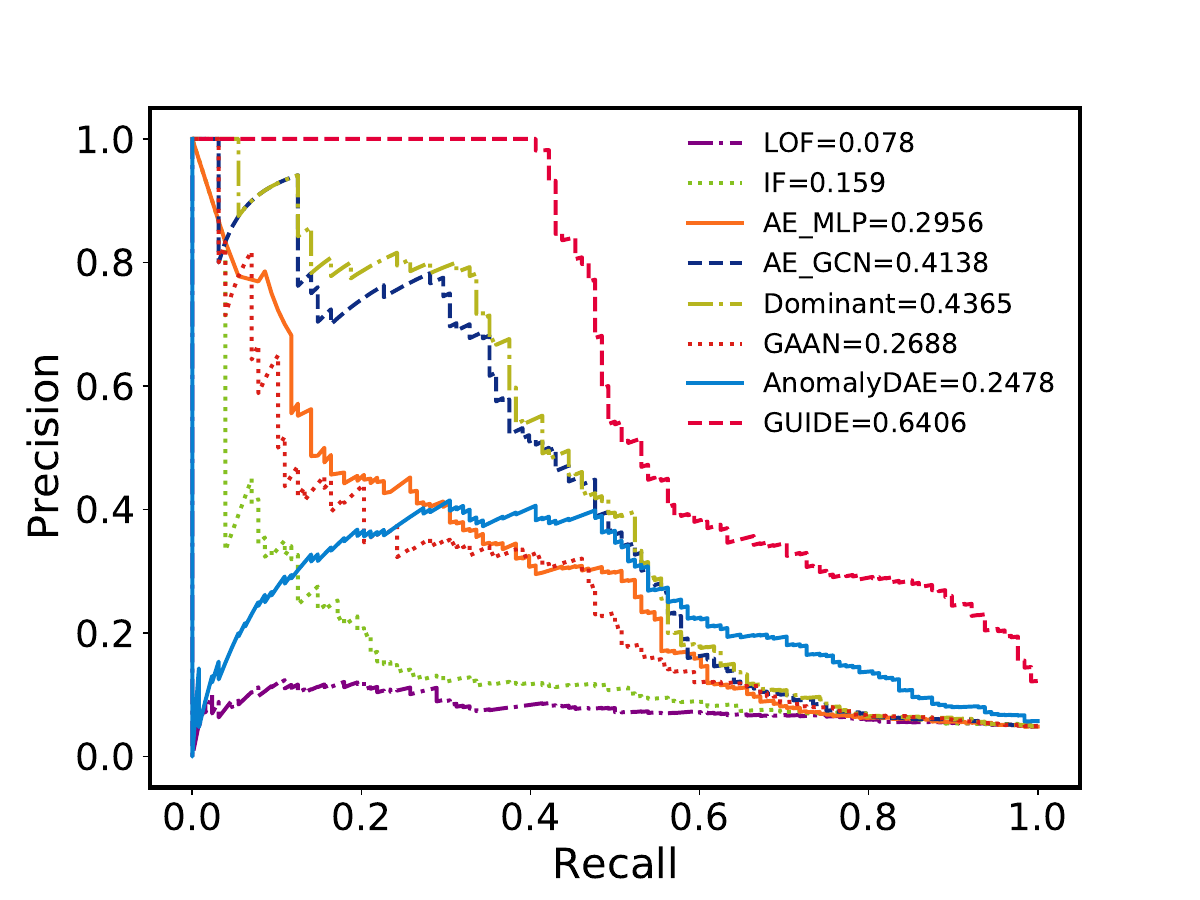}
	}
	\subfigure[DBLP]{
		\includegraphics[scale=0.28]{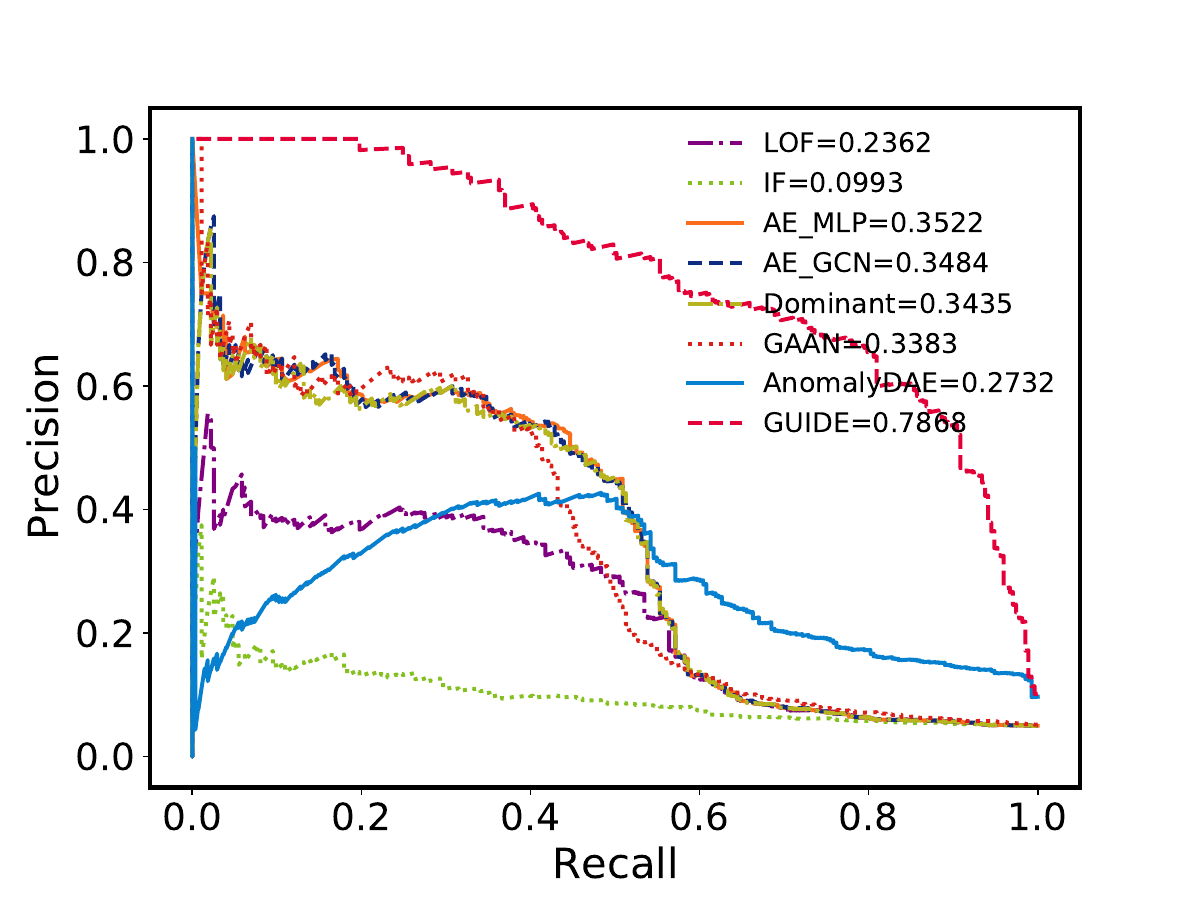}
	}
	\subfigure[Pubmed]{
		\includegraphics[scale=0.28]{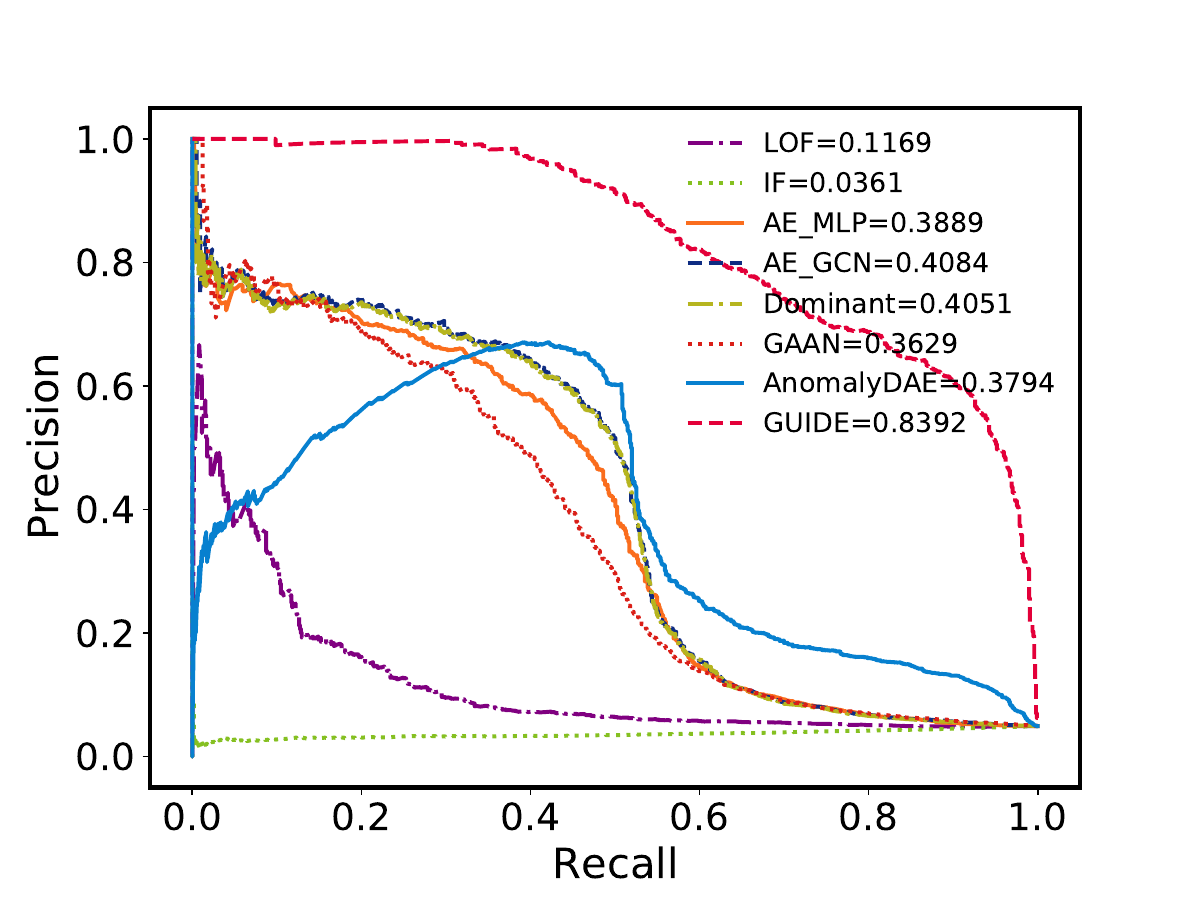}
	}
	\caption{ PR curves and AUC scores of all methods on five real-world datasets.} 
	\label{fig:PR AUC}
\end{figure*}

\begin{itemize} 	
	\item \textbf{ACM\footnote{\url{https://github.com/shenxiaocam/CDNE/blob/master/CDNE_codes/code/CDNE/data}\label{data}}}: ACM is a citation network dataset extracted from the
	Association for Computer Machinery, which is composed of 16,484 scientific publications. And each edge denotes the citation relationship of papers in the network. The attributes of paper consist of sparse bag-of-words features extracted from the paper title.

	\item \textbf{Citation}\textsuperscript{\ref{data}}: Citation is a citation network dataset composed of 8,935 nodes and 15,098 edges, where nodes denote scientific publications, edges represent citation links between publications. The attributes for nodes are the sparse bag-of-words features extracted from the article title.

	\item \textbf{Cora\footnote{\url{https://github.com/kimiyoung/planetoid/tree/master/data}\label{data2}}}: Cora is a citation network dataset including 2,708 nodes, with 5,429 edges to indicate the citation relation. Each node is a scientific publication which is indicated by a binary feature vector.
	
	\item \textbf{DBLP}\textsuperscript{\ref{data}}: DBLP is a citation network dataset composed of 5,484 scientific publications collected from the DBLP Computer Science Bibliography. While the 8,117 edges are the citation relations among different papers. The node attributes are extracted from the article title.
	
	\item \textbf{Pubmed}\textsuperscript{\ref{data2}}: Pubmed contains 19,717 scientific publications with 44,338 links indicating the citation relations between publications. The bag-of-words representations of documents are regarded as the node attributes.
\end{itemize}
Due to the absence of anomalies in the datasets, we refer to two widely used methods \cite{DBLP:conf/wsdm/DingLL19} to inject structural anomalies and attribute anomalies for each dataset, respectively. On one hand, we generate some small cliques to perturb the topological structure of the network. The intuition behind this method is that the nodes in a small clique are much more connected to each other than average, which are always an anomalous structure in many scenarios \cite{DBLP:conf/isi/Skillicorn07}. Therefore, we randomly choose $p$ nodes from the attributed network as a small clique, and make them fully connected. Then all nodes in the small clique are considered as structural anomalies. We perform this process a total of $q$ times and finally generate $q$ small cliques. So there are total $p \times q$ structural anomalies. In experiments, the size of a small clique  $p$ is set to 15. The number of small cliques $q$ is fine-tuned according to different datasets.

On the other hand, to inject attribute anomalies, which have the same number of structural anomalies, we first randomly pick another $p \times q$ nodes. Then we randomly choose another $k$ nodes from the attributed network for each attribute perturbation node $i$ and calculate the Euclidean distance between node $i$ and all the $k$ nodes. Finally, the node $j$ among the $k$ nodes whose Euclidean distance with node $i$ is the largest exchanges the attributes with node $i$. We list the details of these five real-world datasets in Table ~\ref{Table:DS}. 

\subsection{Experimental Settings}
In this section, we describe the compared anomaly detection methods and common evaluation metrics in detail.

\textbf{Baseline Methods.} The GUIDE model is compared with the following baseline methods:
\begin{table*}[]
		\caption{Experimental results of different anomaly detection methods w.r.t. recall@K.}
	\resizebox{1\textwidth}{!}{
		\belowrulesep=0pt    \aboverulesep=0pt
		\begin{tabular}{|l|c|l|l|l|l|l|l|l|l|l|l|l|l|l|l|}
			\midrule[1pt]
			\multicolumn{16}{|c|}{Recall@K}                                                                                                                                \\ \midrule[1pt]
			&
			\multicolumn{3}{c|}{ACM}                                                                  & \multicolumn{3}{c|}{Citation}                                                 & \multicolumn{3}{c|}{Cora}                                                     & \multicolumn{3}{c|}{DBLP}                                                     & \multicolumn{3}{c|}{Pubmed}                                                   \\ \hline
			K          & 50                                  & \multicolumn{1}{c|}{100} & \multicolumn{1}{c|}{150} & \multicolumn{1}{c|}{50} & \multicolumn{1}{c|}{100} & \multicolumn{1}{c|}{150} & \multicolumn{1}{c|}{50} & \multicolumn{1}{c|}{100} & \multicolumn{1}{c|}{150} & \multicolumn{1}{c|}{50} & \multicolumn{1}{c|}{100} & \multicolumn{1}{c|}{150} & \multicolumn{1}{c|}{50} & \multicolumn{1}{c|}{100} & \multicolumn{1}{c|}{150} \\ \midrule[1pt]
			LOF        & \multicolumn{1}{l|}{0.054}          & 0.094                    & 0.132                    & 0.048                   & 0.092                    & 0.134                    & 0.031                   & 0.078                    & 0.125                    & 0.073                   & 0.139                    & 0.205                    & 0.023                   & 0.043                    & 0.063                    \\ \hline
			IF         & \multicolumn{1}{l|}{0.015}          & 0.028                    & 0.037                    & 0.025                   & 0.030                    & 0.046                    & 0.125                   & 0.171                    & 0.211                    & 0.040                   & 0.054                    & 0.088                    & 0.002                   & 0.002                    & 0.003                    \\ \hline
			AE\_MLP        & \multicolumn{1}{l|}{0.074}          & 0.148                    & 0.213                    & 0.078                   & 0.159                    & 0.235                    & 0.180                   & 0.305                    & 0.383                    & 0.109                   & 0.208                    & 0.322                    & 0.038                   & 0.076                    & 0.116                    \\ \hline
			AE\_GCN        & \multicolumn{1}{l|}{0.074}          & 0.145                    & 0.224                    & 0.081                   & 0.166                    & 0.237                    & 0.297                   & 0.398                    & 0.477                    & 0.114                   & 0.209                    & 0.322                    & 0.039                   & 0.076                    & 0.113                    \\ \hline
			Dominant   & \multicolumn{1}{l|}{0.070}          & 0.152                    & 0.222                    & 0.083                   & 0.157                    & 0.237                    & 0.313                   & 0.414                    & 0.484                    & 0.110                   & 0.212                    & 0.315                    & 0.039                   & 0.077                    & 0.112                    \\ \hline
			GAAN       & \multicolumn{1}{l|}{0.071}          & 0.126                    & 0.201                    & 0.078                   & 0.154                    & 0.221                    & 0.164                   & 0.266                    & 0.391                    & 0.117                   & 0.230                    & 0.326                    & 0.040                   & 0.077                    & 0.113                    \\ \hline
			AnomalyDAE & \multicolumn{1}{l|}{0.044}          & 0.118                    & 0.194                    & 0.016                   & 0.060                    & 0.106                    & 0.109                   & 0.313                    & 0.453                    & 0.029                   & 0.095                    & 0.176                    & 0.017                   & 0.038                    & 0.066                    \\ \hline
			GUIDE       & \multicolumn{1}{l|}{\textbf{0.109}} & \textbf{0.217}           & \textbf{0.318}           & \textbf{0.115}          & \textbf{0.230}           & \textbf{0.346}           & \textbf{0.391}          & \textbf{0.484}           & \textbf{0.539}           & \textbf{0.183}          & \textbf{0.341}           & \textbf{0.458}           & \textbf{0.051}          & \textbf{0.102}           & \textbf{0.153}           \\ \hline
			\midrule[0.8pt]
	\end{tabular}}
	\label{Table:recall}
\end{table*}
\begin{itemize} 
	
	\item \textbf{ LOF (Local Outlier Factor)} \cite{DBLP:conf/sigmod/BreunigKNS00} detects anomalies by comparing the local reachability density of the node with their neighbors.

	\item \textbf{ IF (Isolation Forest)} \cite{DBLP:conf/icdm/LiuTZ08} is an attribute based detection method, which detects anomalies utilizing their susceptibility to isolation.
	
	\item \textbf{AE\_MLP (Autoencoder\_Multilayer Perceptron Perceptron)} is a classic neural network model that can reconstruct node attributes to detect anomalies.
	
	\item \textbf{AE\_GCN (Autoencoder\_Graph Convolutional Network)} \cite{DBLP:conf/iclr/KipfW17} is a deep learning model that can detect anomalies by reconstructing node attributes.
	
	\item \textbf{Dominant (Deep Anomaly
		Detection on Attributed Ne-tworks)} \cite{DBLP:conf/sdm/DingLBL19} detects anomalies from both the attribute and structural perspectives by calculating the reconstruction loss of attributes and structures, respectively.
	
	\item \textbf{GAAN (Generative Adversarial Attributed Network Anomaly Detection)} \cite{DBLP:conf/cikm/ChenLWDLB20} is a generative adversarial anomaly detection model, which train jointly reconstruction loss and discriminator loss to detects anomalies. 
	
	\item \textbf{AnomalyDAE (Deep Joint Representation Learning
		Framework for Anomaly Detection through A Dual Autoencoder)} \cite{DBLP:conf/icassp/FanZL20} is a dual autoencoder learning model for anomaly detection. It can learn effectively the complex interactions between node attribute and network structure.
\end{itemize}

\textbf{Evaluation Metrics.} To evaluate the performances of each algorithm, three widely used metrics are adopted to compare different methods in this paper:

\begin{itemize} 
	
	\item \textbf{ROC-AUC:} The ROC curve can measure the relationship between true positive rate (TP) and false positive rate (FP) in different thresholds. AUC value is the area under the ROC curve, which approaches 1 indicating the better performance of the model.
	
	\item \textbf{PR-AUC:}  The PR curve is the curve of precision against recall at different thresholds. AUC value is the area under the PR curve. The higher AUC value means the higher precision and recall.
	
	\item \textbf{Recall@K:}  We utilize Recall@K to calculate the proportion of true anomalous nodes that an anomaly detection method found in all the ground truth anomalies:
	\begin{equation}
	\label{eq:Recall@K}
	Recall@K = \frac{\left| true~anomalies~in~queried~anomalies \right|}{\left| all~true~anomalies \right|}.\nonumber
	\end{equation} 
\end{itemize}
\begin{figure*}[t]
	\centering
	\subfigure[ACM]{
		\includegraphics[scale=0.43]{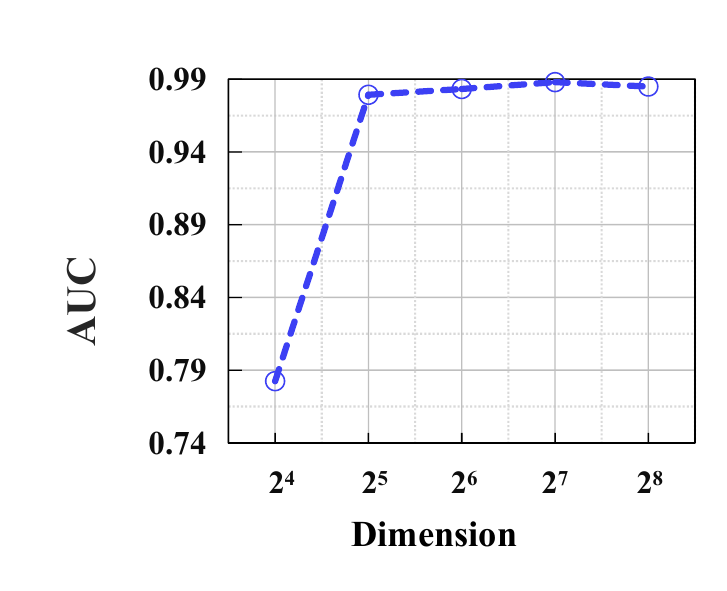}
	}
	\subfigure[Citation]{
		\includegraphics[scale=0.43]{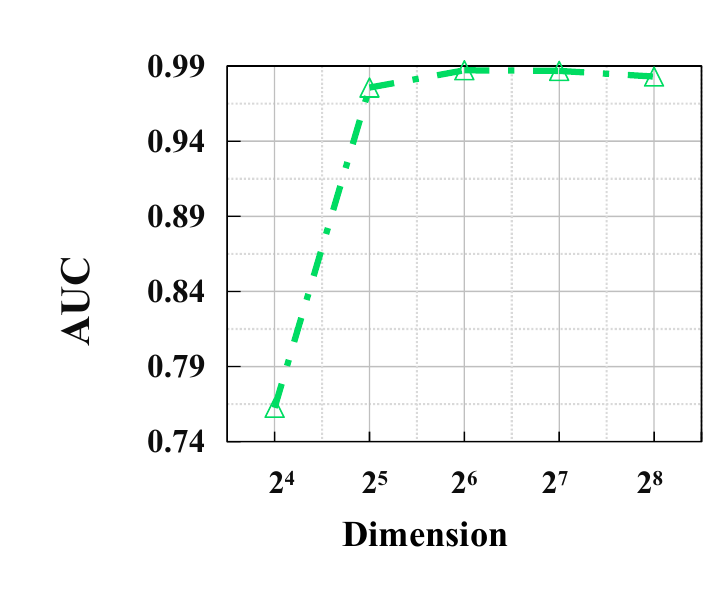}
	}
	\subfigure[Cora]{
		\includegraphics[scale=0.43]{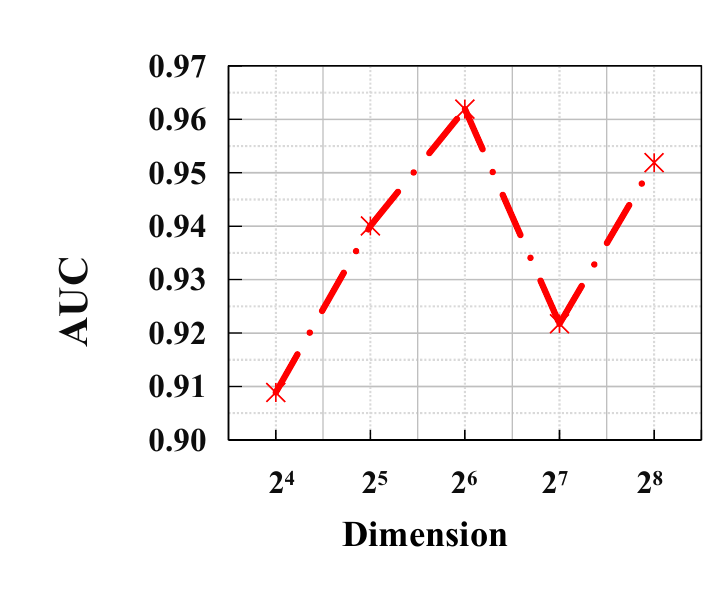}
	}
	\subfigure[DBLP]{
		\includegraphics[scale=0.43]{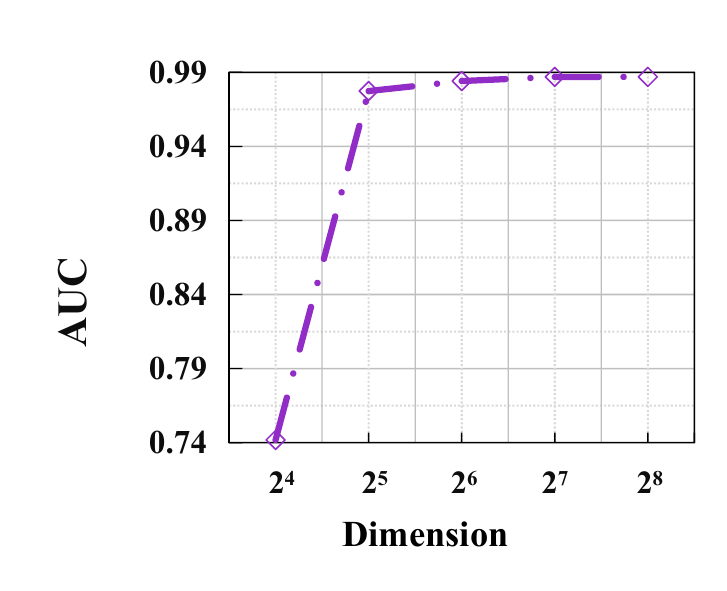}
	}
	\subfigure[Pubmed]{
		\includegraphics[scale=0.43]{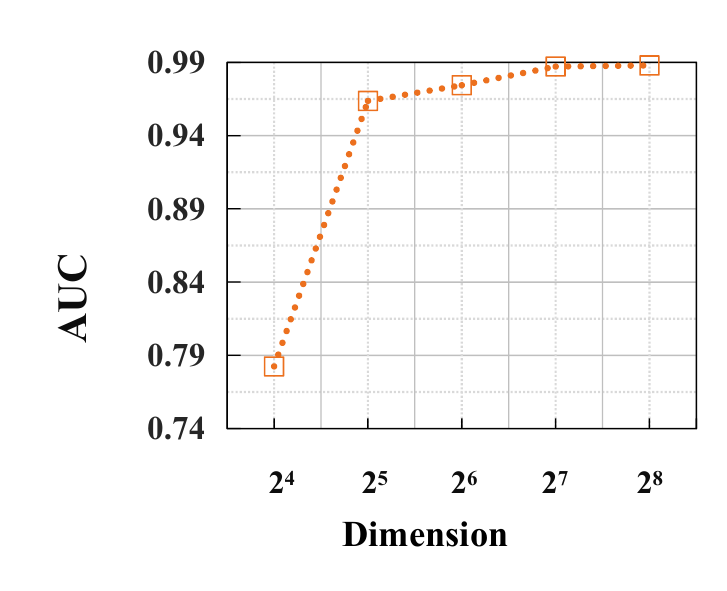}
	}
	
	\caption{Impact of different embedding dimension size w.r.t. AUC values.} 
	\label{fig:embedding dimension}
\end{figure*}

\begin{figure}[htbp]
	\centering
	\includegraphics[width=0.41\textwidth]{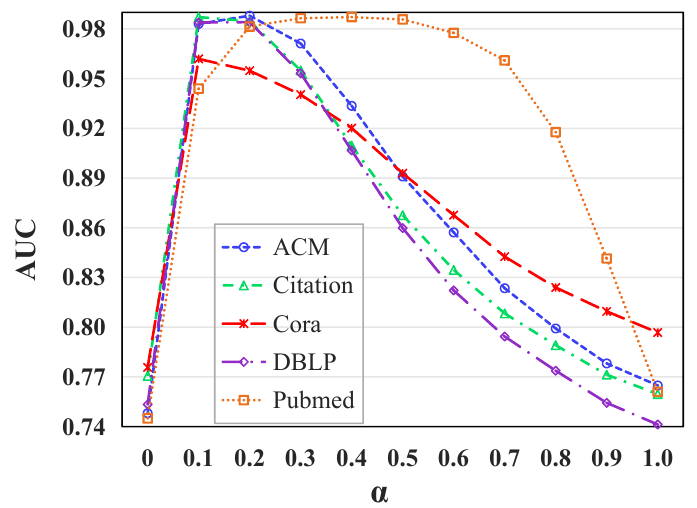}
	\caption{ Impact of different $\alpha$ w.r.t. AUC values.}
	\label{fig:parameter}
\end{figure}

\textbf{Parameter Settings.} In the experiment, the Adam \cite{DBLP:journals/corr/KingmaB14} algorithm is adopted to optimize the loss function on different datasets. We train the GUIDE model for 200 epochs and the learning rate is set to 0.001. We have also optimized the hyperparameters of the model on five real-world datasets through a parameter sensitivity experiment. For other baseline methods, we keep the settings described in the original papers.

\subsection{Experimental Results}
In our experiments, the performance of the GUIDE model is evaluated on multiple metrics by comparing it with the above baseline methods. We first show the experimental results w.r.t. ROC-AUC in Fig \ref{fig:ROC AUC}. While the experimental results in terms of PR-AUC are shown in Fig \ref{fig:PR AUC}. Then the results w.r.t. Recall@K in Table \ref{Table:recall}. According to these experimental results, we obtain the observations as follows:

\begin{itemize}
	\item Our proposed GUIDE model significantly outperforms other baseline methods. The experimental results of the GUIDE model in terms of ROC-AUC, PR-AUC, and Recall@K have a great improvement compared to all baselines. It demonstrates the effectiveness of GUIDE for anomaly detection on attributed networks.
	\item The deep learning models (AE\_MLP, AE\_GCN, Dominant, GAAN, AnomalyDAE, and GUIDE) outperform the conventional anomaly detection methods (LOF and IF). For example, for ROC-AUC on Pubmed dataset, Dominant outperforms LOF by 17.14\% and IF by 39.59\% respectively, AnomalyDAE outperforms LOF by 29.97\% and IF by 52.42\% respectively, and GUIDE outperforms LOF by 40.37\% and IF by 62.82\% respectively. These methods break through the limitations of the shallow mechanism and can effectively address the key challenges of anomaly detection on attributed networks.
	\item AE\_MLP, AE\_GCN, and Dominant have similar performances in terms of ROC-AUC. Compared to Dominant, AE\_MLP and, AE\_GCN only considered reconstructing node attributes but ignored the reconstruction of the first-order structure. It demonstrates that node attributes are more important than the first-order structure of the network for anomaly detection on attributed networks. Therefore, it will not lead to too poor performance when ignoring the first-order structure of the network. This conclusion was also confirmed in \cite{DBLP:conf/sdm/DingLBL19}.
	\item Although AnomalyDAE has shown its superior performance in terms of ROC-AUC, it cannot achieve satisfying results in terms of PR-AUC. For instance, the experimental results of the AnomalyDAE model in terms of ROC-AUC are only lower than those of our proposed model. But its experimental results in terms of PR-AUC are lower than those of most baseline methods. It confirms that AnomalyDAE is not suitable for scenarios with data imbalance. Compared with it, our GUIDE model has reached optimal performance in terms of both ROC-AUC and PR-AUC. Specifically, our ROC-AUC and PR-AUC scores increases by 5.99\% and 42.88\% on ACM, 8.47\% and 55.48\% on Citation, 10.08\% and 39.28\% on Cora, and 8.07\% and 51.36\% on DBLP, and 10.40\% and 45.98\% on Pubmed compared to AnomalyDAE. It shows that the outperformance of GUIDE in scenarios with data imbalance.
	\item Compared with other baseline methods, GUIDE can discover more true anomalous nodes within the ranking list of limited length according to the results of recall@K. Especially, compared with Dominant, our recall@K on Cora increases by 7.8\% in top 50 ranked nodes, 7.0\% in top 100 ranked nodes, and 5.5\% in top 150 ranked nodes. Besides, compared with AnomalyDAE, our recall@K on ACM increases by 6.5\% in top 50 ranked nodes, 9.9\% in top 100 ranked nodes, and 12.4\% in top 150 ranked nodes. It demonstrates the superiority of GUIDE for anomaly detection within the ranking list of limited length.

\end{itemize}

\begin{table}[]
	\caption{ Impact of different structural autoencoder w.r.t. AUC values.}
	\resizebox{0.46\textwidth}{!}{
		\begin{tabular}{|c|ccccc|}
			\hline
			Model       & ACM             & Citation        & Cora            & DBLP            & Pubmed          \\ \hline
			GUIDE\_GCNEN & 0.9784          & 0.9797          & 0.9414          & 0.9531          & 0.9762          \\
			GUIDE\_GCNDE & 0.9727          & 0.9665          & 0.9263          & 0.9534          & 0.9591          \\
			GUIDE\_GCN   & 0.9763          & 0.9596          & 0.9368          & 0.9577          & 0.9501          \\
			GUIDE        & \textbf{0.9879} & \textbf{0.9852} & \textbf{0.9506} & \textbf{0.9790} & \textbf{0.9863} \\ \hline
	\end{tabular}}
	\label{Table:GCN ablation}
\end{table}

\subsection{Ablation Experiment}
To confirm the effectiveness of considering higher-order structures in our framework, we replace the graph node attention network in the structural autoencoder with a graph convolutional network. The specific operations are as follows:
\begin{itemize}
	\item \textbf{GUIDE\_GCNEN:} The structure encoder is replaced with GCN, and the structure decoder is constant.
	\item \textbf{GUIDE\_GCNDE:} The structure encoder is constant, and the structure decoder is replaced with GCN.
	\item \textbf{GUIDE\_GCN:} Both structure encoder and structure decoder are replaced with GCN.
\end{itemize}

We perform evaluations for these methods on five real-world datasets, respectively. The experimental results are shown in Table \ref{Table:GCN ablation}. We found that although the GUIDE model achieved the best performance on all five datasets, the performance do not decrease significantly after replacing the structure autoencoder. It proves that our framework is effective and the higher-order structures play a key role in anomaly detection on attributed networks.
\subsection{Parameter Analysis}

In this section, we study the parameter sensitivity of different numbers of the embedding dimension and the balance parameter $\alpha$ for anomaly detection. The experiment results are presented in Fig. \ref{fig:embedding dimension} and Fig. \ref{fig:parameter}, respectively. We can observe that too low dimension would degrade the performance of the model in Fig. \ref{fig:embedding dimension}, which is caused by underfitting. Moreover, in Fig. \ref{fig:parameter}, we can see that if GUIDE only consider the higher-order structures reconstruction errors $(\alpha=0)$ or the attribute reconstruction errors $(\alpha=1)$, which will lead to poor performance. It demonstrates that the anomaly detection on attributed networks should pay attention to the attributes and the higher-order structures of the node simultaneously. Meanwhile, we find that GUIDE can achieve the best performance when $\alpha$ is around 0.1 to 0.3 on five datasets. It adequately verified the importance of higher-order structures for anomaly detection on attributed networks. 

\section{Conclusion}\label{sec:conclusion}
In this paper, we design a higher-order structure based unsupervised learning framework GUIDE for anomaly detection. Specifically, we employ higher-order network structures when detecting anomalies, and address the limitations of existing methods in modeling complex interaction patterns between multiple entities in the real world. We use dual autoencoders to detect anomalies from both the attribute and higher-order structure perspectives, respectively. To further improve the ability to learn the higher-order structures, we introduce a graph node attention layer, which utilizes the higher-order structure attention mechanism to effectively capture the structural difference between the node and its neighbors. Finally, We can calculate the anomaly score of nodes by the node attribute and higher-order structure reconstruction loss, and sort to find the anomaly. The experiment results on five real-world datasets show that GUIDE outperforms all baseline methods in terms of ROC-AUC, PR-AUC, and Recall@K.

Different higher-order structures generally correspond to various interaction patterns in the real world. For example, the motif M32 can represent the citation relationship between three papers in citation networks. Motif M41 can represent the four-person collaboration relationship in collaboration networks. The significance of different motifs in a certain network should be further evaluated and thus promoting anomaly detection.

\section*{Acknowledgment}
This work is partially supported by National Natural Science Foundation of China under Grant No. 62102060. And the authors would like to thank Chen Cao from Dalian University of Technology for many helpful comments with this manuscript.
\bibliographystyle{ieeetr}
\bibliography{ref}

\end{document}